\pdfoutput=1
\documentclass{article}

    \usepackage[preprint]{neurips_2021}

\usepackage[utf8]{inputenc} % allow utf-8 input
\usepackage[T1]{fontenc}    % use 8-bit T1 fonts
\usepackage{hyperref}       % hyperlinks
\usepackage{url}            % simple URL typesetting
\usepackage{booktabs}       % professional-quality tables
\usepackage{amsfonts}       % blackboard math symbols
\usepackage{nicefrac}       % compact symbols for 1/2, etc.
\usepackage{microtype}      % microtypography
\usepackage{amsmath}
\usepackage{amssymb}
\usepackage{overpic}
\usepackage{pifont}
\usepackage{siunitx}
\usepackage{wrapfig}
\usepackage{graphicx}
\usepackage[numbers]{natbib}

\newcommand{\ie}{\emph{i.e.}}
\newcommand{\eg}{\emph{e.g.}}

\newcommand{\addFig}[1]{}
\newcommand{\addFigs}[1]{}

\renewcommand{\addFig}[1]{\includegraphics[width=0.135\linewidth]{figures/attention_maps/baseline/#1}}
\newcommand{\addFigConv}[1]{\includegraphics[width=0.135\linewidth]{figures/attention_maps/conv_3x3/#1}}

\renewcommand{\addFigs}[1]{\addFig{layer_#1_head_mean_bs_mean.jpg}}
\newcommand{\addFigsConv}[1]{\addFigConv{layer_#1_head_mean_bs_mean.jpg}}

\usepackage{float} 
\usepackage[dvipsnames]{xcolor}
\definecolor{mygreen}{RGB}{0,100,0}
\definecolor{myblue}{RGB}{0,0,240}
\definecolor{myred}{RGB}{200,0,0}

\def \pzo {\phantom{0}} 
\def \dzo {\phantom{00}}

\def \OURS {Refined-ViT}

\hypersetup{colorlinks=true,linkcolor=black,citecolor=mygreen,urlcolor=black}

\title{Distributed Local Attention for Vision Transformers}
\title{Transformer with Convolution Augmented Self-attention}
\title{Global Attention Needs Local Information}
\title{Attention Meets Convolution: Hybrid Attention for Vision Transformer}

\title{High Resolution Global Attention}
\title{Pay Attention to the global context with fine-grained local features}
\title{Global Attention Needs Fined-grained Local Features}
\title{Both Broadness and Details Are Important}
\title{  Diversity Augmented  Self-attention for Vision Transformers }
\title{  Transforming Self-attention for Vision Transformers }
\title{     Attention Augmented   Vision Transformers }

\title{Refiner: Refining Self-attention for \\ Vision Transformers}

\author{Daquan Zhou,
Yujun Shi,
Bingyi Kang,
Weihao Yu, \AND
\vspace{3mm}
Zihang Jiang, 
Yuan Li,
Xiaojie Jin,
Qibin Hou,
Jiashi Feng \\
\textsuperscript{}National University of Singapore\\
\texttt{\small {\{zhoudaquan21, andrewhoux, weihaoyu6, xjjin0731\}}@gmail.com}
\\
\texttt{\small{\{jzihang, kang, shi.yujun\}@u.nus.edu}},
\texttt{\small elefjia@nus.edu.sg}
}

\begin{document}

\maketitle

% \begin{abstract}
%   Vision Transformers (ViTs) have shown compelling performance  in image classification tasks compared with CNNs. Different from CNNs that  locally aggregate  features   via convolutions, ViTs rely on the global-range multi-head  self-attention for feature aggregation, offering high flexibility in modeling long-range dependencies of the local features. However, such flexibility also brings challenges to the training of ViTs, making them demand larger-scale  datasets for  (pre-)training than CNNs to achieve similar performance. In this work, we dig deeper into  this prominent issue    and uncover it largely attributes to  an inherent limitation of the global  self-attention: the attention maps tend to be overly smoothed and thus the aggregated features become   unfavorably similar to each other.  To address this, we propose two solutions: attention expansion that implicitly increases  the number of attention heads to diversify the attention maps and a novel convolutional attention mechanism, which together give a new    \nameofmethod{} method. Significantly, our proposed  \shortnameofmethod{}  enables ViTs to achieve 85.73\% top-1 classification accuracy on the ImageNet benchmark with less than 100M parameters. 
% \end{abstract}

\begin{abstract}
  Vision Transformers (ViTs) have shown competitive accuracy  in image classification tasks  compared with CNNs. Yet, they generally require much more data for model pre-training.  Most of recent works thus are dedicated to designing more complex architectures or training methods to address the data-efficiency issue of ViTs. However,  few of them explore   improving the self-attention mechanism, a key factor distinguishing ViTs from   CNNs. 
  % In this work, we look into this negelected 
  %demonstrate that such low data-efficiency of ViTs largely attributes to the over-smoothing issue of their multi-head self-attention maps. 
  Different from existing works, we introduce a conceptually simple scheme, called \emph{refiner}, to  directly refine the self-attention maps of ViTs.
  % Aiming at addressing such prominent limitation of ViTs, we investigate two 
  Specifically, refiner  explores   \textit{attention expansion} that projects the multi-head attention maps to a higher-dimensional space to promote  their diversity. Further,  refiner    applies convolutions to   
    augment  local patterns of the attention maps, which we show   is equivalent to a \textit{distributed local attention}\textemdash features are  aggregated locally  with learnable kernels  and then  globally aggregated with self-attention.   Extensive experiments demonstrate that refiner works  surprisingly well. Significantly, it  enables  ViTs to achieve 86\% top-1 classification accuracy on   ImageNet  with only 81M parameters. Code is publicly available at \url{https://github.com/zhoudaquan/Refiner_ViT}.
\end{abstract}

\section{Introduction}
\label{sec:intro}
 
% \begin{wrapfigure}[20]{r}{0.4\textwidth}
%     \centering
%     \vspace{-10mm}
%     \includegraphics[width=0.4\textwidth]{figures/atten_visualize.pdf}
%     \vspace{-3mm}
%     \caption{\small Visualization of the attention maps from our method (left) and vanilla self-attention (SA) of ViTs (right). SA tends to give uniform attention to all the locations while our proposed DLA presents  more precise focus on the target object.}
%     \label{fig:atten_visualize}
% \end{wrapfigure}

Recent progress on image classification   largely attributes to  the development of  vision transformers (ViTs) \cite{dosovitskiy2020image, touvron2020training, touvron2021going}. Unlike convolutional neural networks (CNNs) that rely on  convolutions  (\eg, using $3\times 3$ kernels) to process     features locally~\cite{tan2019efficientnet, he2016deep, sandler2018mobilenetv2, zhou2019neural, xie2017aggregated}, %hard  convolutional \emph{spatial inductive bias},  \ie, the locality  and weight sharing,
ViTs take advantage of the self-attention (SA) mechanism~\cite{vaswani2017attention} 
to establish the global-range relation among the image patch features (\textit{a.k.a.} tokens) and aggregate them across all the spatial locations. Such global information aggregation nature of SA      substantially increases ViT's expressiveness and has been proven to be more flexible than convolutions \cite{srinivas2021bottleneck, touvron2020training, yuan2021tokens, jiang2021token, liu2021swin}
on  learning   image representations. 
However, it also leads ViTs to need  extraordinarily larger amount of data for pre-training (\eg, JFT-300M) or much longer training time and stronger data augmentations than   its  CNN counterparts to achieve similar performance \cite{touvron2020training}. 
%However, it also leads the ViTs to need much larger amount of pre-training data (\eg,   JFT-300M) or longer training time than   its  CNN counterparts to achieve similar performance.   

% Thereby, ViTs can achieve better classification results than CNNs with similar model size in the data-plentiful regime~\cite{dosovitskiy2020image}.  \footnote{The data efficient training methods proposed in \cite{touvron2020training} needs advanced training recipes and well trained teacher models. Otherwise, the large amount of data is still essential for SOTA performance.}

\begin{figure}[t]
    \centering
    \small
    \includegraphics[width=1.0\linewidth]{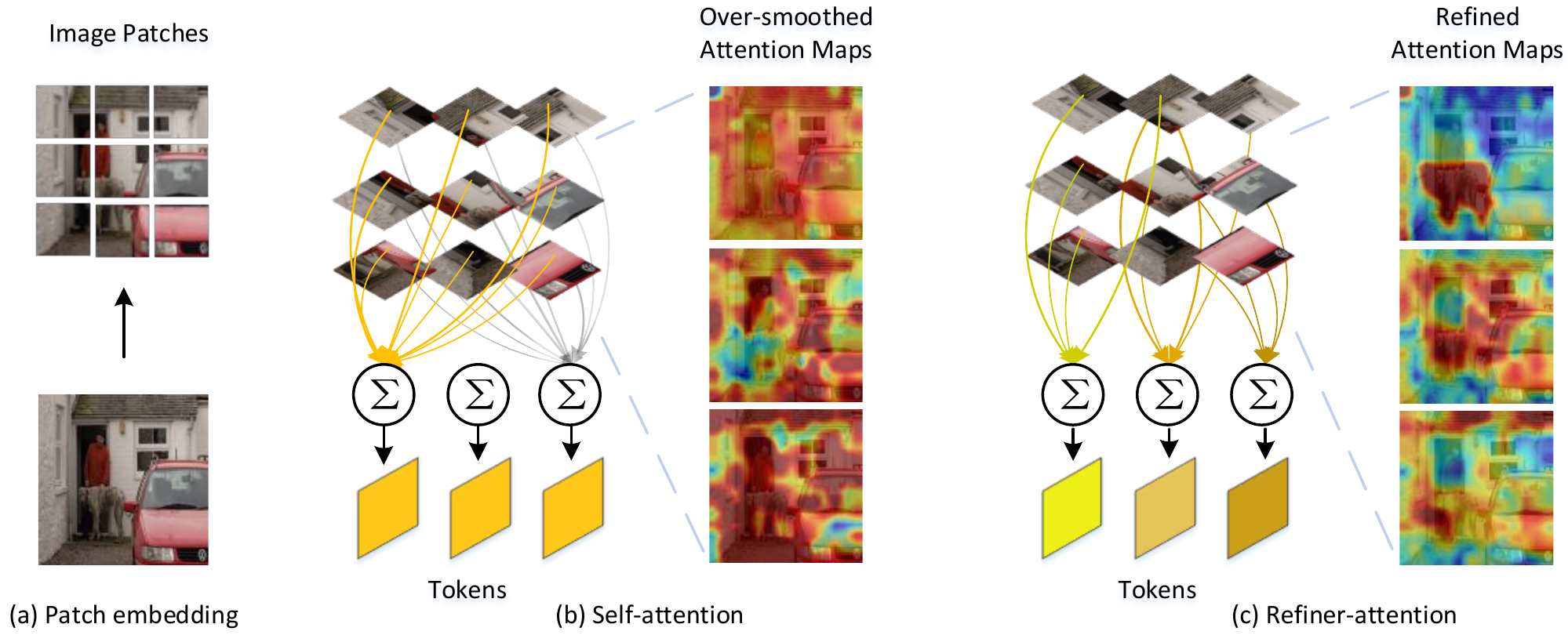}
    \caption{Illustration on our motivation. (a) The input image is regularly partitioned into patches for patch embedding. (b) The token-wise attention maps from vanilla self-attention of ViTs tend to be uniform, and thus they aggregate  all the patch embeddings densely and generate overly-similar tokens. (c) Differently, our proposed refiner augments the attention maps into diverse ones with enhanced  local patterns, such that they aggregate the  token features more selectively and the resulting tokens are distinguishable from each other. }
    \label{fig:overall_flow_comp}
    \vspace{-4mm}
\end{figure}

 Most of recent works thus  design  more complex architectures or training methods to improve data-efficiency of ViTs~\cite{touvron2020training,yuan2021tokens,srinivas2021bottleneck}. However,  few of them pay attention to the SA component of ViTs.
 Within a SA block in ViTs, each token is updated by aggregating the features from \emph{all} tokens according to the attention maps, as shown in Fig.~(\ref{fig:overall_flow_comp})(b). In this way, the tokens of a layer are able to sufficiently exchange information with each other and thus offer great expressiveness. However, this can also cause different tokens to become more and more similar, especially as the model goes deeper (more information aggregation). 
%  Within the SA block of a ViT, each token aggregates the  features of \emph{all} the tokens  according to the attention maps,   as shown in Fig.~(\ref{fig:overall_flow_comp})(b). Despite of offering great expressiveness, such operation may mix  the features of irrelevant tokens (\eg, the irrelevant object tokens and background tokens) together, thus   leading the learned embedding of each token to become indistinguishable from the others when the model goes deeper. 
 This phenomenon, also known as over-smoothing, has been identified  by some recent studies~\cite{zhou2021deepvit, touvron2021going,dong2021attention} to   largely degrade the  performance of the  ViTs. 
%\textemdash the features of different tokens become overly similar to each other with limited   discriminativeness among them. 
%Such phenomenon  is also observed by some recent studies , constituting  
% Such limitation   arguably hurts       data-efficiency of the original ViT \cite{dosovitskiy2020image}, and 
   
In this work, we explore to address the above limitation by directly refining their self-attention maps. 
% In this work, we explore to address the above limitation and improve the data efficiency and performance of ViTs by directly refining their self-attention maps.
Specifically, inspired by   the recent  works~\cite{touvron2021going} demonstrating that increasing the number of SA heads can effectively   improve the model performance, we investigate how to   promote the  diversity of the attention maps using a similar strategy. However, given  a transformer model with fixed embedding dimension, directly increasing the number of heads will reduce the  number of embeddings allocated to each head, making the computed attention map less comprehensive and accurate as shown in Tab. 9 in \cite{touvron2021going}. To address this dilemma,   we explore  \emph{attention expansion}   that linearly projects  the multi-head attention maps   to a higher-dimensional space spanned by a larger number of attention maps. As such, the number of attention heads (used for computing the attention maps) are  implicitly increased  without reducing   the embedding  dimension per head, enabling the model to enjoy   both benefits from more SA heads and high embedding dimension.

Additionally, we argue that ignoring the local relationship among the tokens is another main cause of the above mentioned over-smoothing issue of global SA. The locality (local receptive fields) and spatial invariance (weight sharing)  have been proven to be the key of the success of CNNs across many computer vision tasks \cite{li2017scale, sohn2012learning, wang2017residual, he2016deep, bello2021lambdanetworks}. Therefore, we explore how to leverage the  convolution  to augment the attention mechanism of ViTs. Some recent works have studied injecting convolutional locality  into ViTs, \eg,  integrating both convolution and global self-attention into a hybrid model~\cite{dosovitskiy2020image, srinivas2021bottleneck}. However, they mostly consider directly applying convolution to the token features, keeping the convolutional block and the SA block separately. 
Differently,   we explore introducing convolution to the attention maps directly to augment their spatial-context and local patterns, thus increasing their diversity. Such an approach, as we explain later, can be understood as a \textit{distributed local attention} mechanism that combines both strengths of self-attention (global-range modeling) and convolution (reinforcing local patterns). 

Based on the above two strategies, we introduce the  \textit{refiner} scheme to directly enhance self-attention maps within ViTs. Though being conceptually simple,  it works surprisingly well.  As shown in Fig. \ref{fig:overall_flow_comp}(c), the attention map with refiner has  preciser focus on the target object, compared to the vanilla one. By directly applying refiner to the SA module, the performance of the ViT-Base \cite{dosovitskiy2020image} is improved by 1.7\% on ImageNet under the same training recipes with negligible memory overhead (less than 1M). 
We have experimentally verified that this improvement is consistent when using more complex training recipes \cite{touvron2020training, touvron2021going} and the results are shown in Appendix.
Surprisingly, refiner is also applicable for natural language processing tasks.  On GLUE \cite{wang2018glue}, it increases the average score  by 1\% over the state-of-the-art BERT~\cite{devlin2018bert}. % With advanced augmentation and training recipes \cite{touvron2020training}, DLA achieves 85.73\% top-1 accuracy on ImageNet with only 81M parameters. This is the new state-of-the-art (SOTA) accuracy under 100M number of parameters. % \td{add some impressive experiment results. Discussion on Figure 1. }
 
In summary, our contributions are: 
% \begin{itemize}
First, we tackle the important low data-efficiency issue of ViTs from a new angle, \ie, directly refining the self-attention maps. We  investigate a simple yet novel refiner scheme and find it   works surprisingly well, even outperforming the state-of-the-art methods using sophisticated training recipes or leveraging extra models.  Secondly, the introduced refiner  makes  several interesting  modifications on the self-attention mechanism that would be inspiring for future works. Its attention expansion gives a new viewpoint for   addressing the common dilemma on the trade-off between the number of  attention heads and the embedding dimensions. Besides, different form the common practice  that   applies convolutions to features, it proves that convolutions can  effectively augment  attention maps as well  and    improve  the model performance  significantly.

\section{Related Works}
\label{sec:related_works}
\paragraph{Transformers for vision tasks} Transformers~\cite{vaswani2017attention}
were originally developed for solving natural language processing tasks and achieved remarkable success \cite{brown2020language, radford2018improving}. Different from other deep neural network architectures like CNNs and RNNs \cite{liu2016recurrent, sak2014long}, transformers rely on the self-attention mechanism to perform the global interactions among features instead of relying on certain inductive biases (\eg, locality and weight sharing from CNNs). This grants transformers stronger capacity in learning from large-scale data. Those properties of transformers motivate researchers to explore its application on computer vision tasks, including image enhancement~\cite{chen2020pre,yang2020learning}, image classification~\cite{dosovitskiy2020image,yuan2021tokens,zhou2021deepvit}, object detection~\cite{carion2020end,zhu2020deformable,dai2020up,zheng2020end}, segmentation~\cite{wang2020end,liu2021swin, zheng2020end},  video processing~\cite{zeng2020learning,zhou2018end} and 3D point cloud data processing~\cite{zhao2020point, guo2020pct}.   Among them, Vision Transformer (ViT)~\cite{dosovitskiy2020image} is one of the early pure-transformer based models that achieve state-of-the-art performance on ImageNet. However, due to the larger model capacity, ViTs need extraordinarily large-scale datasets (\eg, ImageNet-22K and JFT-300M) for pre-training to achieve comparable performance with CNNs at similar model size. Some follow-up models thus try to resolve the low data-efficiency limitation either by modifying the model architecture~\cite{yuan2021tokens,touvron2020training,touvron2021going} or adopting new training techniques like knowledge distillation~\cite{touvron2020training}.

\vspace{-3mm}
\paragraph{Injecting locality into ViTs}
% Most of   previous   works use self-attention as an auxiliary mechanism to augment CNNs on computer vision tasks \cite{hutt2005analysis}. Differently, 
Recent   works find that injecting the convolution locality into transformer blocks can boost performance of ViTs. As initially proposed in \cite{dosovitskiy2020image, srinivas2021bottleneck}, the   hybrid ViT  uses   pre-trained CNN models to extract the patch embedding from the input images and then deploys multiple transformer blocks  for feature processing.  
%Recent research works attempt to combine the advantages of transformers and CNNs but still treat them as separate operators. 
\cite{li2021localvit}  replaces the feed forward block in the transformer block with an inverted residual one \cite{sandler2018mobilenetv2, howard2017mobilenets} and \cite{peng2021conformer} propose to build two concurrent branches: one with convolution and the other one with transformer. However, they all leverage convolutions to strengthen the features. Besides, the interaction between the global information aggregation at the SA block and the local information extraction in the convolution block is not well explored. In this paper, we focus on enhancing the attention maps and explore how the global context and the local context can be combined in a novel way.

\section{Refiner}
\label{sec:method}
% We first review the global self-attention mechanism of ViT. Then we point out the learning difficulties incurred by such attention mechanism    via experiments. We thus propose a new distributed local attention  to remedy the inherent limitation of ViT. 

\subsection{Preliminaries on ViT}
The vision transformer (ViT) first tokenizes an input image by regularly partitioning it into a sequence of $n$ small patches. Each patch is then projected into an embedding vector $x_i $ via a linear projection layer,  with an additional learnable position embedding.   

Then ViT applies multiple multi-head self-attention (MHSA) and feedforward layers to process the  patch embeddings to model their long-range dependencies and evolve the token embedding features. Suppose the input tensor is $X\in\mathbb{R}^{ d_{in} \times n }$, the MHSA applies linear transformation with parameters $W_K, W_Q, W_V$ to embed them into the key $K=W_K X \in \mathbb{R}^{d\times n}$, query $Q=W_Q X \in \mathbb{R}^{d\times n}$ and value $V = W_V X \in \mathbb{R}^{d\times n}$ respectively. Suppose there are $H$ self-attention heads. These embeddings are uniformly split into $H$ segments $Q_h,K_h,V_h \in \mathbb{R}^{ d/H \times n }$. Then the MHSA module computes the head-specific attention matrix (map)\footnote{We interchangeably use attention matrix and attention map to call $A$ when  the context is clear. Specifically,  the attention matrix   means the original $A$ while   the attention map means  reshaped $A$  to the input  size. } $A$ and aggregate the token value features as follows:
\begin{equation}
\label{eqn:vanilla_sa}
    \mathrm{Attention}(X,h) =  A^h V^\top_h  \text{ with } A^h = \mathrm{Softmax}\left(\frac{Q_h^\top K_h}{\sqrt{d/H}}\right),  h = 1, \ldots, H.
\end{equation}

\paragraph{Limitations of ViTs} 
% \td{discuss   these results.}
The ViT  purely relies on the global self-attention to establish the relationship among all the tokens without deploying any spatial structure priors (or inductive biases). Though benefiting in learning flexible features from large-scale  samples, such global-range attention may lead to overly-smoothed attention maps and token features as pointed out by~\cite{zhou2021deepvit,gong2021improve}. We hypothesize this may slow   the feature evolving speed of   ViTs, \ie, the features  change slowly when traversing the model blocks, and make the model perform inferior to the ones with faster feature evolving speed. To verify this, we present a pilot experiment to compare  the feature learning speed of ViTs with similar-size CNNs (ResNet)~\cite{he2016deep}  that incorporate strong locality regularization and DeiT~\cite{touvron2020training} which is a better-performing transformer model trained with   sophisticated   scheme. 
 
\begin{wrapfigure}{r}{0.55\linewidth}
    \centering
    \tiny
    \begin{overpic}[width=0.55\textwidth]{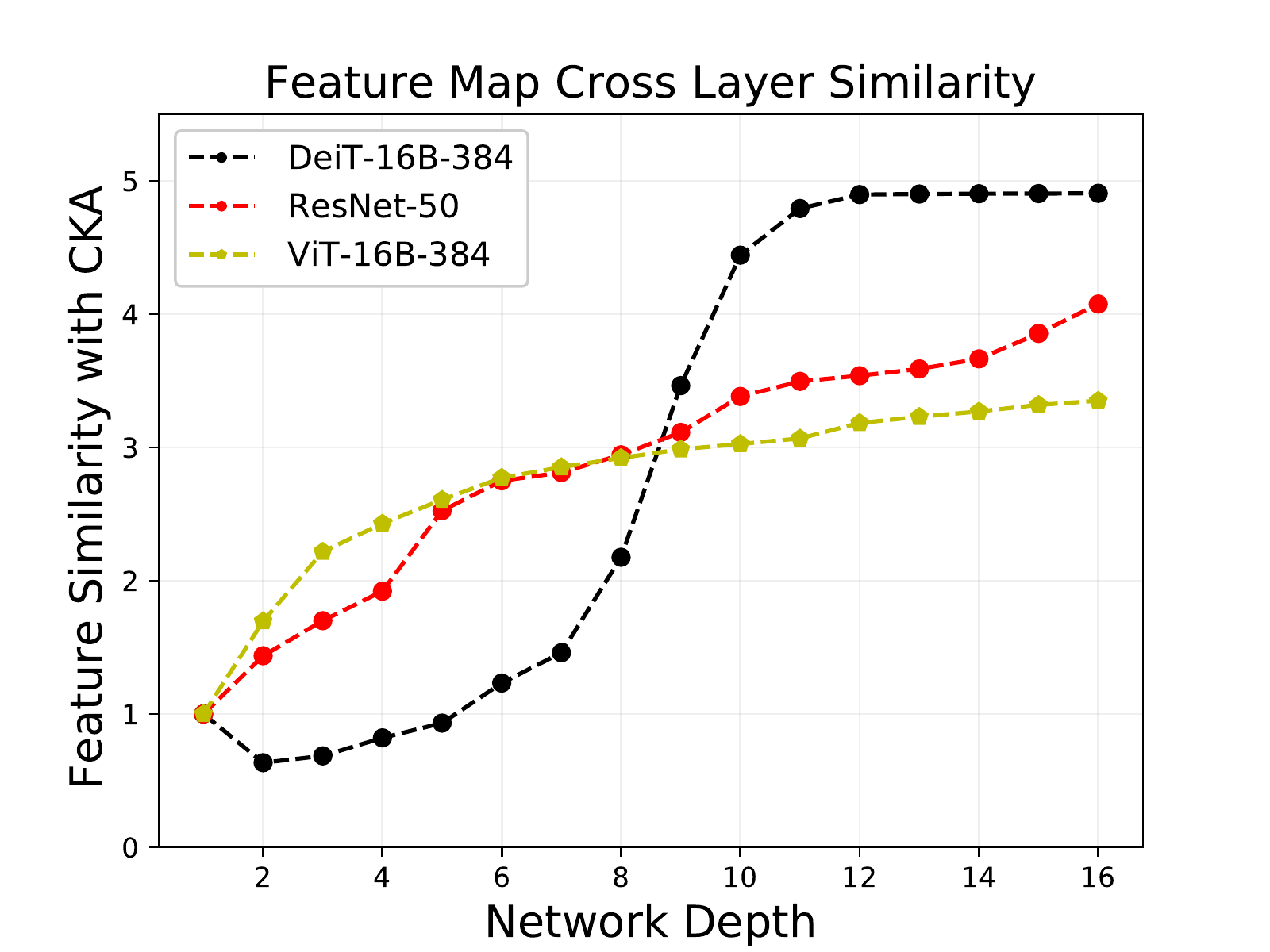}
    \put(40.5,14.5){
        % \tinysize
        \setlength\tabcolsep{0.5mm}
        \renewcommand{\arraystretch}{1}
        \begin{tabular}{l|ccc} 
          Method  & \#Param.  & \#Blocks & Acc.  \\ \hline
          DeiT-16B-384  & 24M & 16 & 80.8\%  \\
          ResNet-50  & 25M & 16 & 80.5\%   \\
          ViT-16B-384  & 24M & 16 & 78.9\%  \\
        \end{tabular}
    }
    \end{overpic}
    % \begin{minipage}[c]{1\textwidth}
    \caption{
    The features of ViT evolves slower than ResNet \cite{he2016deep} and DeiT \cite{touvron2020training} across the model blocks. 
    }
    % \end{minipage}
    % \vspace{-15pt}
	\label{fig:feature_evo}
	\vspace{-4mm}
\end{wrapfigure}
% \end{figure}
In Fig. \ref{fig:feature_evo}, we plot the CKA similarity~\cite{kornblith2019similarity} (ref. its formula to Appendix) between the intermediate token features  at the output of each block and the final output, normalized by the similarity between the first and final layer features. Such a metric captures two properties: how fast the intermediate features converge to the final features and how much the final features are different from the first layer features.  It is clearly shown that for the ViT, the feature evolves slowly after the first several blocks.  Differently, the token features  of ResNet and DeiT keep  evolving faster when the model goes deeper and the final features are more different than the first layer features compared with the ViT. In this work, we investigate whether such limitation of ViTs can be solved by  refining  the attention maps into more diverse and locality-aware ones. 
%
% We find that self-attention typically leads to over-similar token features when the model goes deeper. The average similarity between the tokens of a ViT model with 16 blocks are shown in Fig. \ref{fig:feature_evolving}(b).

\subsection{Attention Expansion}
The MHSA of transformers captures different aspects of the input features~\cite{vaswani2017attention} by projecting them into different subspaces and computing self-attention therein. Thus, increasing the number of attention heads within MHSA, which is shown to be   effective at improving the model performance \cite{touvron2021going}, can potentially increase  diversity among the attention maps. However, as mentioned above, if naively adding more SA heads, for a model with fixed hidden dimension, it is difficult to trade-off the benefit of having more SA heads and the harm of reducing the embedding dimension per head.
% (\ie, the subspace dimension). 

To resolve this issue,    instead of directly expanding the attention heads, we investigate expanding the number of self-attention maps via linear transformation.  Concretely, we use a linear projection $W_A \in \mathbb{R}^{H'\times H}$ with $H' > H$ to project   the attention maps $\mathbf{A}=[A^1; \ldots; A^{H}]$ to a new set of attention maps $\tilde{\mathbf{A}}=[\tilde{A}^1,\ldots, \tilde{A}^{H'}]$ with $\tilde{A}^h = \sum_{i=1}^H W_A(h,i) \cdot A^i, h=1,\ldots,H'$. Then we use the new attention maps $\tilde{\mathbf{A}}$ to aggregate features as in Eqn.~\eqref{eqn:vanilla_sa}.

The above attention expansion implicitly increases the number of SA heads but does not sacrifice the embedding dimensions. To understand  how it works, recall each of the original SA map $A_h$ is computed by $A_h = Q_h^\top K_h$. Here we omit the softmax and normalizing factor for illustration simplicity. We use $w_i$ to denote $W_A(h,i)$.  Thus, the up-projected   attention map is computed by $\tilde{A}^h = \sum_{i=1}^H w_i \cdot Q_i^\top K_i = [w_1\cdot Q_1, \ldots, w_H \cdot Q_H]^\top K$. Different from the vanilla MHSA that divides $Q,K$ into $H$ sub-matrices for SA computation, the attention expansion    reweighs the   $Q$ features with learnable scalars  and uses the complete reweighed $Q$ to compute the SA maps. Getting rid of the feature division, the attention expansion effectively solves the above issue. When only one weight scalar is non-zero, it   degenerates to the vanilla MHSA. The attention expansion is similar to the talking-heads attention~\cite{shazeer2020talking,zhou2021deepvit} which are proven effective in improving   attention diversity but differently, it expands the  number of attention maps while talking-heads does not change the number. 

% \td{add explanation on where to add the ReLU. } A ReLU non-linearity is applied onto the new attention maps, followed by another linear projection for dimension adjustment:
% \begin{equation*}
%     {\mathbf{A}}' = \mathrm{ReLU} (\tilde{\mathbf{A}} * w) \times  P'.
% \end{equation*}

 \begin{figure}[t]
    \centering
    \includegraphics[width=1\linewidth]{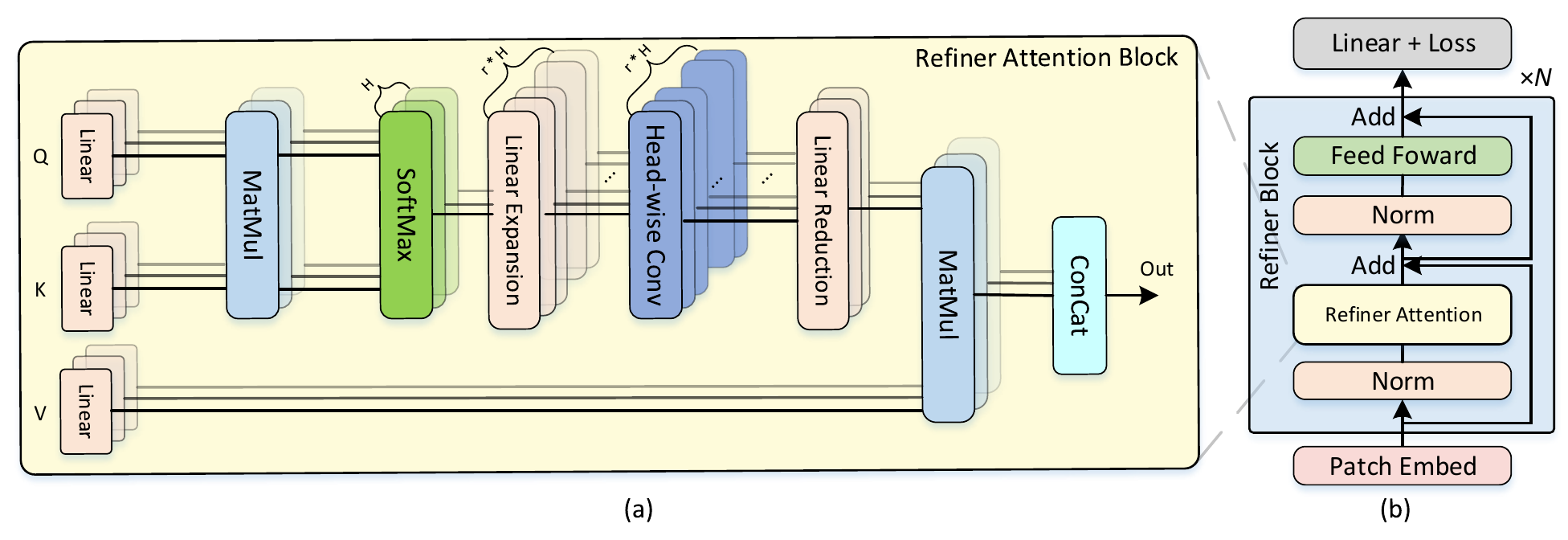}
    \caption{(a)  Architecture design of refiner. Different from the vanilla self-attention block, the refiner applies linear attention expansion to attention maps output from the softmax operation to increase their number. Then head-wise spatial convolution is applied to augment these expanded attention maps. Finally another linear projection is deployed to reduce the number of attention maps to the original one. Note that $r = H'/H$ is the expansion ratio. (b)  Modified transformer block with refiner as a drop-in component. }
    \label{fig:model_arch}
\end{figure}

\subsection{Distributed Local Attention}

Our second strategy is to investigate how convolutions can help enhance  self-attention maps since convolutions are good at strengthening  local patterns.  Different from some earlier trials on introducing convolutions to process features   in ViTs~\cite{srinivas2021bottleneck,li2021localvit,peng2021conformer}, we explore   directly applying convolutions to the self-attention maps. 

Given the pre-computed self-attention matrix $A^h$ from head $h$, we  apply a learnable convolution kernel $\mathbf{w} \in \mathbb{R}^{k\times k}$ of size $k\times k$ to modulate it  as follows: 
\begin{equation*}
     {A}^{h^{*}}_{i,j}  = \sum_{a,b=1}^k \mathbf{w}_{a,b} \cdot A^h_{i- \lfloor \frac{k}{2} \rfloor +a,j-\lfloor \frac{k}{2}\rfloor +b}.
\end{equation*}
Then the standard SA-based feature aggregation as Eqn.~\eqref{eqn:vanilla_sa} is conducted      with this new attention matrix $\tilde{\mathbf{v}}_i  = \sum_{j=1}^n  \tilde{A}^{*}_{i,j} \cdot  \mathbf{v}_j$.

Though being conceptually simple, the above operation establishes an interesting synergy  between the global context aggregation (with self-attention) and local context modeling (with convolution).
To see this,   consider applying 1D convolution $\mathbf{w}$ of length $k$ to obtain  the convolution-augmented SA matrix, with which the feature aggregation becomes 
\begin{equation*}
\begin{aligned}
    \tilde{\mathbf{v}}_i 
    & = \sum_{j=1}^n A^{h^{*}}_{i,j} \cdot  \mathbf{v}_j  = \sum_{j=1}^n \left( \sum_{a=1}^k \mathbf{w}_{a} \cdot A^h_{i,j-\lfloor \frac{k}{2}\rfloor +a} \right) \cdot  \mathbf{v}_j  \\
    & = \sum_{j=1}^n \left( \sum_{a=1}^k \mathbf{w}_{a} \cdot A^h_{i,j-\lfloor \frac{k}{2}\rfloor +a}  \cdot  \mathbf{v}_j \right) 
      = \sum_{j=1}^n   \left( \sum_{a=1}^k A^h_{i,j} \cdot \mathbf{w}_{a} \cdot \mathbf{v}_{j-a+\lfloor \frac{k}{2}\rfloor} \right). 
    \end{aligned}
\end{equation*} 
The above clearly shows that the feature aggregation based on the convolution-processed  attention matrix  is equivalent to: (1) applying the convolution $\mathbf{w}$, with a location-specific reweighing scalar $A^h_{i,j}$,  to aggregate features locally  at first and (2) summing over the locally aggregated features. Therefore, we name such an operation as \emph{distributed local attention} (DLA).

We empirically find DLA works pretty well in the  experiments. We conjecture that DLA effectively combines   strengths of the  two popular feature aggregation schemes. Convolutional schemes are strong at capturing   local patterns but inefficient in global-range modeling. Moreover,  stacking multiple  convolutions  may suffer the increasingly center-biased   receptive  field~\cite{luo2017understanding}, leading the model to ignore the  features at the image boundary. In contrast, self-attention based feature aggregation can effectively model the global-range relation among features but suffer the over-smoothing issue. DLA can effectively overcome their limitations and better model the local and global context jointly.

\subsection{Refiner and Refined-ViT} 

The attention expansion and distributed local attention can be easily implemented via $1\times 1$ and $3 \times 3$ convolutions.  
Fig.~\ref{fig:model_arch} illustrates the refiner architecture. Different from the vanilla self-attention, refiner expands the attention maps at first and then applies head-wise convolution  to process the maps.  Refiner  additionally introduces the attention reduction for  reducing the model computation cost and maintaining  the model embedding dimension. Specifically, similar to attention expansion, the attention reduction applies   linear projection to reduce the number of attention maps from $H'$ to the original $H$, after the distributed local attention. Refiner can serve as a drop-in module to directly replace the vanilla self-attention   in each ViT transformer block, giving the \textit{Refined-ViT} model.

\section{Experiments}
\label{sec:exp}
%  We   aiming to address the following questions: (1) whether it can effective  improve performance of ViTs by augmenting the attention maps? (2) how its performance will be affected by different design choices on the attention expansion and local attention kernels? (3) is it   applicable   for  improving  transformers for other tasks beyond ViTs? 
% \vspace{-4mm}

\subsection{Experiment Setup}

% \subsubsection{Datasets}
We mainly evaluate Refiner for the image classification task. Besides, we also conduct experiments on natural language processing tasks to investigate its generalizability to NLP transformer models. 
\vspace{-3mm}
\paragraph{Computer vision} We evaluate the effectiveness of the refiner on ImageNet \cite{deng2009imagenet}. For a fair comparison with other methods, we first replace the SA module with refiner on ViT-Base \cite{dosovitskiy2020image} model as it is the most frequently used one \cite{dosovitskiy2020image, touvron2020training, yuan2020revisiting, touvron2021going}. When comparing with other state-of-the-art methods, we modify the number of transformer blocks and the embedding dimensions to increase the efficiency in the same manner as \cite{yuan2021tokens, zhou2021deepvit}. 
\vspace{-3mm}
\paragraph{Natural language processing} We evaluate our model on the General Language Understanding Evaluation (GLUE) benchmark~\cite{wang2018glue}. GLUE benchmark includes various tasks which are formatted as single sentence or sentence pair classification. See Appendix for more details of all tasks.
We measure accuracy for MNLI, QNLI, QQP, RTE, SST, Spearman correlation for STS and Matthews correlation for CoLA. 
The GLUE score is the average   score for all the 8 tasks. 

\vspace{-3mm}
\paragraph{Training setup} All the experiments are   conducted upon PyTorch~\cite{paszke2019pytorch} and the timm~\cite{rw2019timm} library. The models are trained on ImageNet-1k from scratch without auxiliary dataset.
For the ablation experiments, we follow the standard training schedule and train our models on the ImageNet dataset for 300 epochs. When compared to state-of-the-art (SOTA) models, we use the advanced training recipes as proposed in \cite{touvron2020training}. Detailed training hyper-parameters   are listed in the appendix.

\subsection{Analysis}
\label{exp:analysis}
Refiner introduces attention expansion and convolutional local kernels to augment the self-attention maps of ViTs. Here we individually investigate their effectiveness through ablative studies. 

%=====================================

\begin{table}[h]
    \footnotesize
    \setlength\tabcolsep{2.1pt}
    \centering
    \caption{\small  Effect of the expansion ratio in  attention expansion, which varies from 1 to 6. The model is \OURS{} with 16 blocks, 12 attention heads and 384 embedding dimension. }
% \vspace{-2mm}
    \label{tab:attn_expansion_ratio}
    % \begin{tabular}{lccccccc} \toprule
    % Model & \#Blocks  & Embedding &Expansion Ratio & Params & Converge  (\#Epoch) &  Top-1 (\%)\\ \midrule
    % \OURS &  16 & 384  & 1 & 25M & 300 & 82.3\\ 
    % \OURS &  16 & 384  & 2 & 25M & 300 & 82.8\\ 
    % \OURS &  16 & 384  & 3 & 25M & 273 & 82.8\\
    % \OURS &  16 & 384  & 4 & 25M & 270 & 82.9\\
    % \OURS &  16 & 384  & 6 & 25M & 261 & 83.0\\ \bottomrule
    % \end{tabular}
    %     \begin{tabular}{ccccc} \toprule
    % Expan. Ratio & Params & Converge  (\#Epoch) &  Top-1 (\%)\\ \midrule
    %  1 & 25M & 300 & 82.3\\ 
    % 2 & 25M & 300 & 82.8\\ 
    % 3 & 25M & 273 & 82.8\\
    %  4 & 25M & 270 & 82.9\\
    % 6 & 25M & 261 & 83.0\\ \bottomrule
    % \end{tabular}
       \begin{tabular}{cccc} \toprule
    Expan. Ratio & Params & Converge  (\#Epoch) &  Top-1 (\%)\\ \midrule
     1 & 25M & 300 & 82.3\\ 
    2 & 25M & 300 & 82.8\\ 
    3 & 25M & 273 & 82.8\\
     4 & 25M & 270 & 82.9\\
    6 & 25M & 261 & 83.0\\ \bottomrule
    \end{tabular}
% \vspace{-4mm}
\end{table}
%=====================================

% \begin{table}[h]
%     \footnotesize
%     \setlength\tabcolsep{2.1pt}
%     \centering
%     \caption{Ablation on the impact of the number of the heads for the ViT with 16 blocks and 384 embedding dimension. }
% % \vspace{-2mm}
%     \label{tab:attn_expansion_ratio}
%     % \begin{tabular}{lccccccc} \toprule
%     % Model & \#Blocks  & Embedding &Expansion Ratio & Params & Converge  (\#Epoch) &  Top-1 (\%)\\ \midrule
\paragraph{Effect of attention   expansion} We adopt $1\times 1$ convolution to adjust the number of attention maps. In this ablative study, we vary the expansion ratio $r = H'/H$ from 1 to 6. From the results     in Tab.~\ref{tab:attn_expansion_ratio}, we observe that along with    increased expansion ratio (and more attention maps), the model performance   monotonically increases from 82.3\% to 83.0\%. This clearly demonstrates the effectiveness of attention expansion in improving the model performance. Note when the expansion ratio equals to 1, the attention expansion degenerates to the talking-heads attention \cite{shazeer2020talking,zhou2021deepvit}. Compared with this strong baseline, increasing the ratio to 2 brings 0.5\% top-1 accuracy improvement. Interestingly, using larger expansion ratio speeds up the convergence of model training. Using an expansion ratio of 6 saves the number of training epochs by nearly 13\% (261 vs.\ 300 epochs). This reflects that attention expansion can help   the  model learn the discriminative token features faster. However, when the expansion ratio is larger than 3, the benefits in the model accuracy attenuates. This motivates us to explore the distributed local attention for further performance enhancement.
%     % \OURS &  16 & 384  & 1 & 25M & 300 & 82.3\\ 
%     % \OURS &  16 & 384  & 2 & 25M & 300 & 82.8\\ 
%     % \OURS &  16 & 384  & 3 & 25M & 273 & 82.8\\
%     % \OURS &  16 & 384  & 4 & 25M & 270 & 82.9\\
%     % \OURS &  16 & 384  & 6 & 25M & 261 & 83.0\\ \bottomrule
%     % \end{tabular}
%         \begin{tabular}{ccccc} \toprule
%     \# Heads & ViT-Small Top-1 (\%) &  Refined-ViT Top-1 (\%)\\ \midrule
%      8  & 79.9 & 82.5\\ 
%     12 & 80.0 & 82.6\\ 
%     16 &  80.0 & 82.8\\
%     \bottomrule
%     \end{tabular}
% % \vspace{-4mm}
% \end{table}

% \begin{wraptable}{r}{4.cm}
% \vspace{-2mm}
%   \centering
%   \small
%   \setlength\tabcolsep{1mm}
%   \renewcommand\arraystretch{1}
%   \caption{\small Effect of attention reduction on Refined-ViT-16B with 384 hidden dimension.}
%   \label{tab:attn_reduction}
%   \begin{tabular}{cc} \toprule[0.5pt]
%     Attn. map  &Top-1 (\%)\\ \midrule[0.5pt]  
%       w/o reduction & 82.99
%     \\
%      w/ reduction & 82.95  \\
%     \bottomrule
%   \end{tabular}
% \end{wraptable}

% \vspace{-3mm}
\vspace{-2mm}
\paragraph{Effect of attention   reduction} Refiner  deploys another $1\times 1$ convolution to reduce the number of attention maps from $H'$ to the original number $H$,  after the attention expansion and distributed local attention, in order to reduce the computation overhead due to expansion. We conduct experiments to study whether reduction will hurt model performance.  As observed from Tab.~\ref{tab:attn_reduction}, attention reduction   drops the accuracy very marginally, implying the attention maps have been sufficiently augmented in the higher-dimension space. 

\vspace{-2mm}
\paragraph{Effect of distributed local attention} We then evaluate whether the distributed local attention (DLA)    works consistently well across a broad spectrum of model architectures. We evaluate its effectiveness for various ViTs with 12 to 32 SA blocks, without the attention expansion. Following the common practice~\cite{dosovitskiy2020image}, the hidden dimension is 768 for the ViT-12B and 384 for all the other ViTs.   We set the local attention window size as $k=3$ and use the DLA to replace all the self-attention block within ViT. From the results   summarized in Tab.~\ref{tab:convolution_effect}, DLA  can consistently boost the top-1 accuracy of various ViTs by 1.2\% to 1.7\% with negligible model size increase. Such significant performance boost clearly demonstrates  its effectively and the benefits brought by its ability in  jointly modeling the local and global context of input features. The combined effects of attention expansion and DLA are shown in Tab. \ref{tab:sota}.

\begin{table}[t]
\begin{minipage}[t]{0.55\textwidth}
    \caption{\small  Impacts of convolution on attention maps. We directly apply the $3 \times 3$ convolution on the attention maps from the multi-head self-attention of ViTs with respect to various architectures. We can observe clear improvement for all ViT variants when adding
    the proposed DLA.}
    \vspace{-2mm} 
    \label{tab:convolution_effect}
    \footnotesize
    \setlength\tabcolsep{2.1pt}
    \begin{center}
     \begin{tabular}{l c c c c c c c} 
    \toprule
     Model & \#Blocks & Hidden dim &  \#Heads  & Params &  Top-1  (\%)\\ 
    \midrule
      \textsc{ViT}&  12 & 768 & 12 &   86M & 79.5 \\ 
    %   \textsc{ConvBERTsmall}&+BNK, +SDConv& MLM  & 256 & 64 & 2 & 1  & 14M &75.9\\ 
    ~ +  DLA & 12  & 768 & 12    & 86M & 81.2 \\ 
    \midrule
      \textsc{ViT}&  16 & 384 & 12   & 24M & 78.9\\ 
    %   \textsc{ConvBERTsmall}&+BNK, +SDConv& MLM  & 256 & 64 & 2 & 1  & 14M &75.9\\ 
    ~  +  DLA & 16  & 384 & 12     & 24M & 80.3\\ 
     \midrule
      \textsc{ViT}&  24 & 384 & 12   & 36M & 79.3\\ 
    %   \textsc{ConvBERTsmall}&+BNK, +SDConv& MLM  & 256 & 64 & 2 & 1  & 14M &75.9\\ 
    ~  +  DLA & 24  & 384 & 12   & 36M & 80.9\\ 
     \midrule
      \textsc{ViT}&  32 & 384 & 12   & 48M & 79.2\\ 
    %   \textsc{ConvBERTsmall}&+BNK, +SDConv& MLM  & 256 & 64 & 2 & 1  & 14M &75.9\\ 
    ~  +  DLA & 32  & 384 & 12     & 48M & 81.1 \\ 
    \bottomrule
    \end{tabular}
    % \vspace{-4mm}
    \end{center}
\end{minipage}
\hfill
\begin{minipage}[t]{0.42\linewidth}
    \begin{minipage}[t]{\linewidth}
    \centering
    \small
    \setlength\tabcolsep{0.5mm}
    \renewcommand\arraystretch{1}
    \caption{\small Effect of attention reduction on Refined-ViT-16B with 384 hidden dimension.}
    \label{tab:attn_reduction}
    \vspace{-3pt}
    \begin{tabular}{ccc} \toprule
    Model & Attn. map  &Top-1 (\%)\\ \midrule[0.5pt]  
    Refined-ViT-16B & w/o reduction & 82.99 \\
    Refined-ViT-16B & w/ reduction  & 82.95 \\
    \bottomrule
  \end{tabular}
    \vspace{15.3pt}
    \end{minipage}
    
    \begin{minipage}[b]{\linewidth}
    \caption{\small Evaluation on how the  spatial span within DLA  affects the model performance. We compare the model performance with three different constraints on the local  kernels. }
    \vspace{-8pt}
    \label{tab:spatial_span_effect}
    \footnotesize
    \setlength\tabcolsep{4pt}
    \begin{center}
    \begin{tabular}{ccc}  \toprule
    Model & Constraints       &  Top-1 (\%) \\ \midrule
    Refined-ViT-16B & None    &   83.0\\ 
    Refined-ViT-16B & Spatial  & 82.7\\ 
    Refined-ViT-16B & Row+Col & 81.7 \\ \bottomrule
    \end{tabular}
    % \vspace{-4mm}
    \end{center}
    \end{minipage}
    
\end{minipage}
\end{table}

% \vspace{-2mm}
\begin{wrapfigure}[18]{r}{0.5\textwidth}
    \centering
    \vspace{-4mm}
    \includegraphics[width=0.5\textwidth]{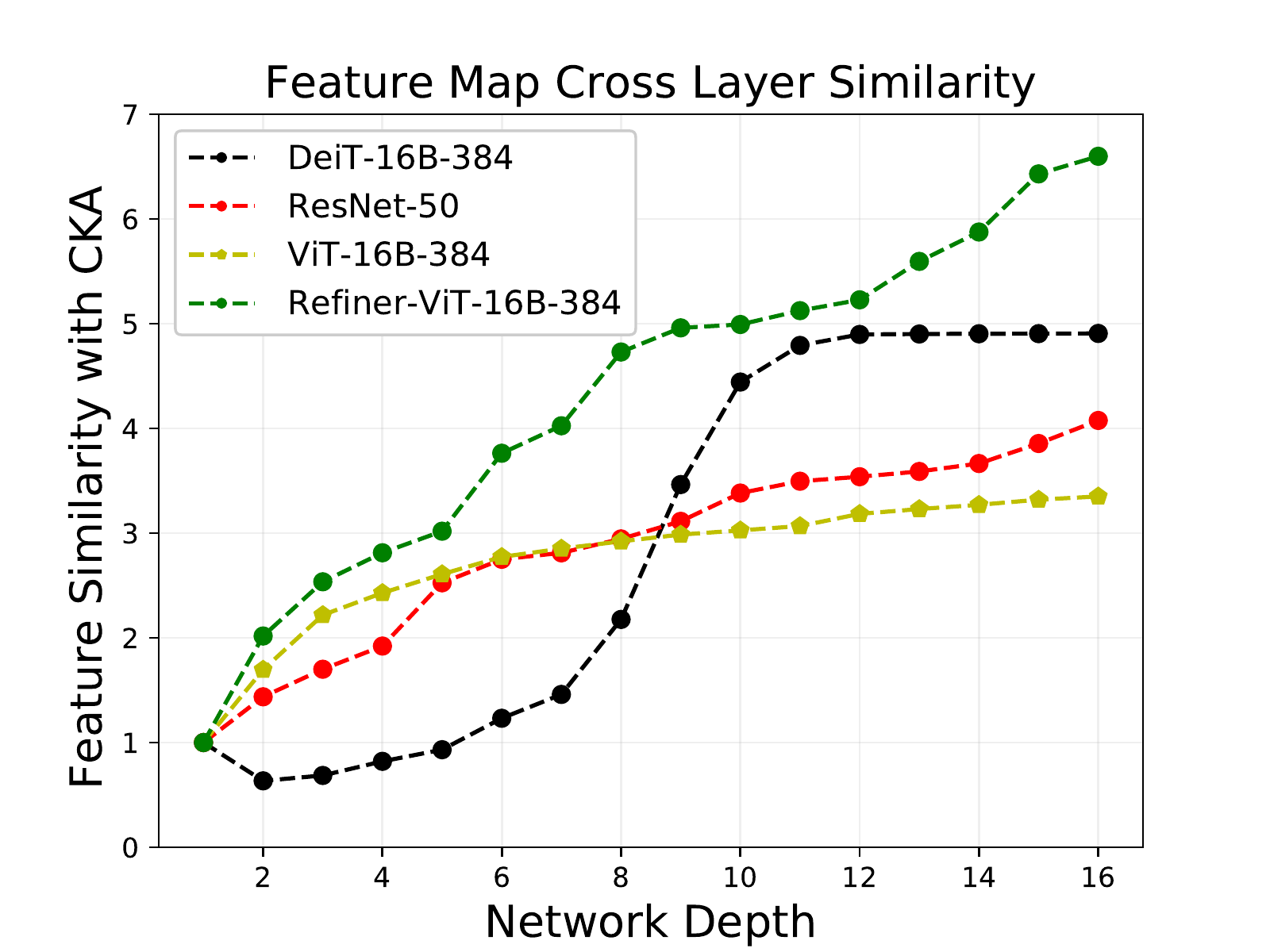}
    \vspace{-4mm}
    \caption{\small Refiner accelerates feature evolving compared with CNNs, the vanilla ViT and the Deit trained with a more complex   scheme.}
    \label{fig:feat_evo_conv}
\end{wrapfigure}

\vspace{-2mm}
\paragraph{Effect of the local attention kernels }  
Refiner directly applies  
the $3\times 3$ convolution   onto the attention maps $A$.
We compare this natural choice with another two feasible choices for the local kernels. The first one is to  apply spatially adjacent convolution on the  reshaped attention maps such that  the aggregated tokens by the local kernels  are spatially adjacent in the original input. Specifically,  
we reshape each row of $A$ into a $\sqrt{n} \times \sqrt{n}$ matrix\footnote{Recall $n$ is the number of tokens and the input image is divided into $\sqrt{n} \times \sqrt{n}$ patches.}  which together form a  $\sqrt{n} \times \sqrt{n} \times n $ tensor and apply $3\times 3$ convolution.  
The second one is to apply the combination of row and column 1D convolution. From the results in Tab.~\ref{tab:spatial_span_effect},   we find the spatially adjacent convolution will slightly hurt the performance while only adopting the 1D convolution   leads to 1.3\% accuracy drop. Directly applying convolution to the attention maps (as what refiner does) gives the best performance, demonstrating the effectiveness of augmenting the local patterns of the attention maps.

% \begin{figure*}[t]
%     \centering
%     \small
%     \setlength\tabcolsep{0.5mm}
%     \begin{overpic}[width=1\linewidth]{}
%         % \put(5, 5){Tokens with CutMix \cite{yun2019cutmix}}
%     \end{overpic}
%     \caption{Ablation on the effects of kernel size.}
%     \label{fig:kernel_size}
% \end{figure*}

% \vspace{-3mm}

% \begin{table}[h]
%   \centering
%   \small
%   \setlength\tabcolsep{1mm}
%   \renewcommand\arraystretch{1}
%   \caption{Ablation on the performance gain under different training recipes.}
%   \label{tab:shared_attn}
%   \begin{tabular}{lcclcc} \toprule[0.5pt]
%     Model & Throughput~(\#Samples\SI[per-mode = symbol]{}{\per\s}) & \#Shared Attention Maps &  Top-1 Acc. (\%)\\ \midrule[0.5pt] \midrule[0.5pt]
%      \OURS-12B-768 &  &  0 & 83.10
%     \\
%     \OURS-12B-768 &  & 1 & 83.04  \\
%     \OURS-12B-768 &  & 2 &  \\
%   \end{tabular}
% \end{table}

\begin{figure}[h]
  \centering
  \footnotesize
  \setlength\tabcolsep{0.2mm}
  \renewcommand\arraystretch{1.0}
  \begin{tabular}{cccccccc}
    \rotatebox{90}{~~~~~~SA} & 
    \addFigs{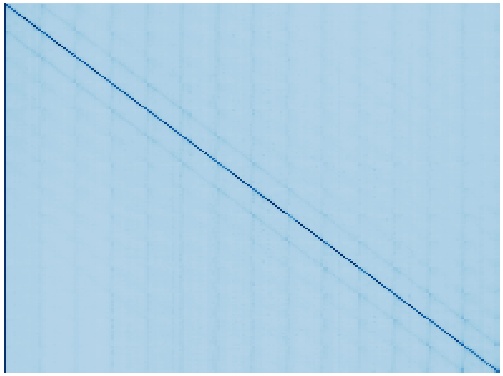} & 
    \addFigs{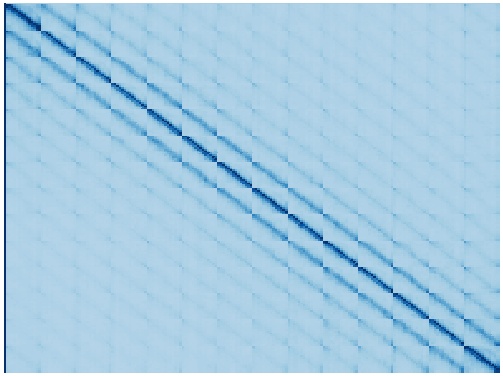} & 
    \addFigs{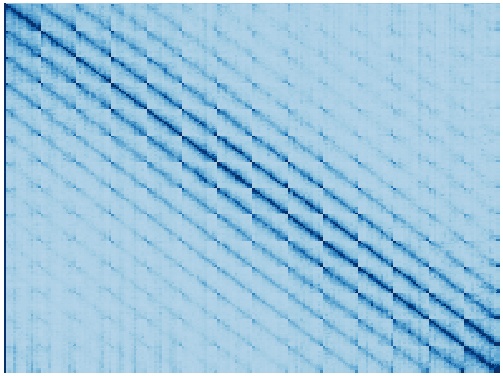} & 
    \addFigs{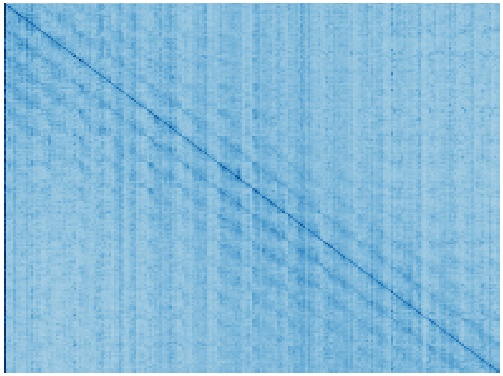} & 
    \addFigs{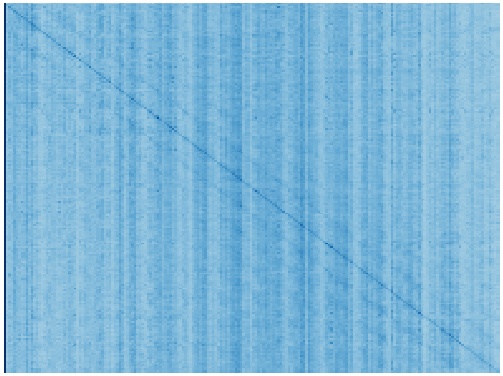} &
    \addFigs{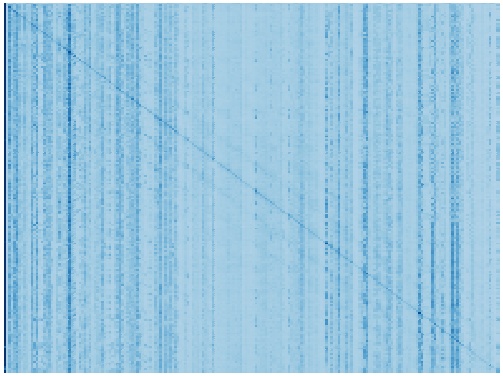} &
    % \addFigs{5} & 
    % \addFigs{6} & 
    % \addFigs{7} & 
    % \addFigs{8} &
    \addFigs{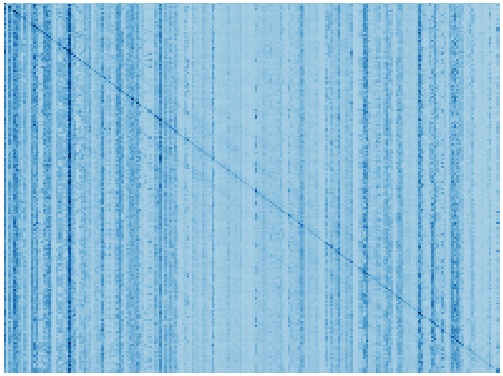} \\
    \rotatebox{90}{~~~Refiner} & 
    \addFigsConv{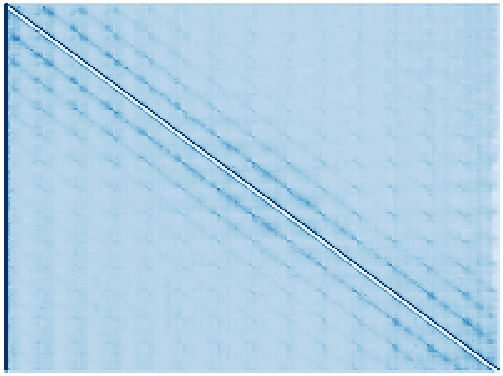} & 
    \addFigsConv{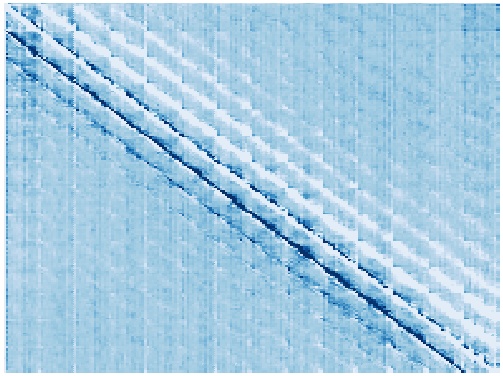} &
    \addFigsConv{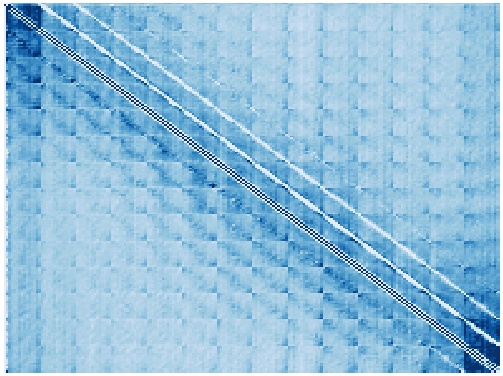} & 
    \addFigsConv{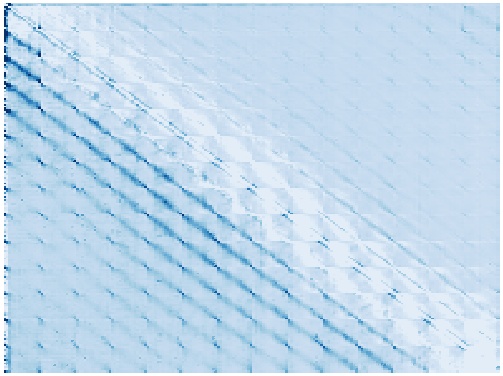} & 
    \addFigsConv{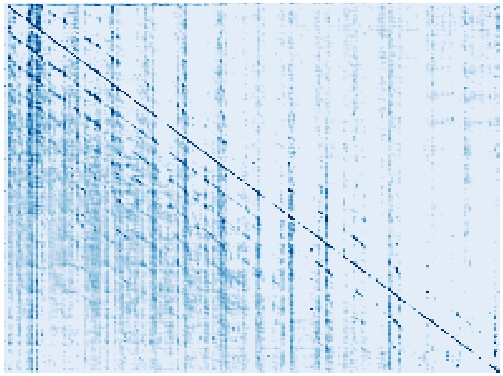} &
    \addFigsConv{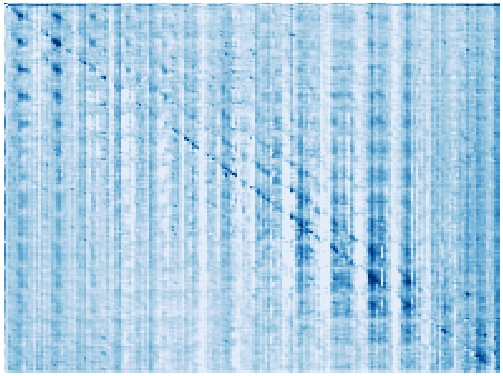} &
    % \addFigs{15} & 
    % \addFigs{16} & 
    % \addFigs{17} & 
    % \addFigs{18} &
    \addFigsConv{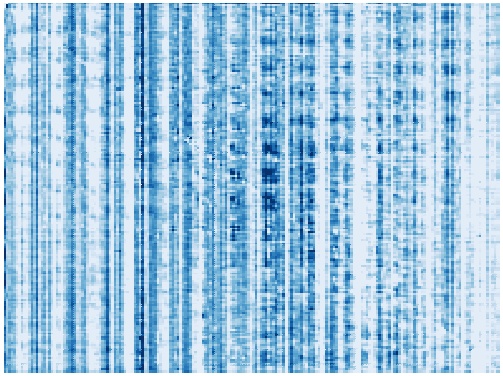} \\
     & Block 1 & Block 4 & Block 11 & Block 18  &Block 23 & Block 29 
     & Block 30
     \\
  \end{tabular}
  \vspace{3pt}
  \caption{\small  Compared with the attention matrices $A$ from the vanilla SA (top), for   deeper blocks, refiner (bottom) strengthens the local patterns of their attention maps,   making them less uniform and better model   local context. 
  }
  \label{fig:vis_atten_map}
  %\vspace{-10pt}
\end{figure}

% \vspace{-3mm}
\paragraph{Refiner augments attention maps and accelerates feature evolving} We visualize the attention maps output from different self-attention blocks without and with the refiner in Fig.~\ref{fig:vis_atten_map}. Clearly, the attention maps after Refiner become not so uniform and present stronger local patterns. We also plot the feature evolving speed in Fig.~\ref{fig:feat_evo_conv},  in terms of the CKA similarity change w.r.t.\ the first block features and Refined-ViT evolves the features much faster than ViTs.

\vspace{-2mm} 
\subsection{Comparison with SOTA}
\label{exp:vision_sota}
We compare refined-ViT with   state-of-the-art models in Tab.~\ref{tab:sota}. For small-sized model, with  224$\times$224  test resolution, Refined-ViT-S (with only 25M parameters) achieves 83.6\% accuracy on ImageNet, which outperforms the strong baseline DeiT-S by   3.7\%. For medium-sized model with 384$\times$ 384   test resolution, the model Refined-ViT-M  achieves the accuracy of 85.6\%, which outperforms the latest SOTA  CaiT-S36 by 0.2\% with 13\% less of parameters (55M vs.\ 68M) and outperforms LV-ViT-M by 1.6\%. Note that CaiT uses knowledge distillation based method to improve their model, requiring much more computations. 
For large-sized model finetuned with input resolution of 448$\times$ 448, refined-ViT achieves 86\%, better than CaiT-M36 with 70\% less of parameters (81M vs.\ 271M). This is the new state-of-the-art results on ImageNet with less than 100M    parameters. Notably, as shown in Tab. \ref{tab:sota}, our models significantly outperform the recent  models that introduce   locality into the token features across all the model sizes. This clearly  demonstrates: first, refining the attention maps (and thus the feature aggregation mechanism) is more effective than augmenting the features for ViTs; secondly,   jointly modeling the local and global context (from DLA in the refiner)  is better than using    the convolution and self-attention separately and deploying them into different blocks.

\subsection{Receptive field calibration}
In this subsection, we present and investigate a simple approach that we find can steadily improve classification performance of  ViT. The approach is a generic one that also works well for CNNs. 

As pointed out in \cite{touvron2019fixing}, pre-processing the training images and testing images separately could lead to a distribution shift between the training and testing regimes. This issue has been named by \cite{touvron2019fixing} as the mismatch between the region of classification (RoC),
%
% the mismatch between the region of classification (RoC) during training and testing phases 
which could degrade   performance of the classification model. 
A common remedy solution is to apply random cropping to get a centrally cropped image patch for classification,  with a certain cropping ratio (less than 1),  at the testing phase. 
% set a cropping ratio at the testing phase to get a central cropped image patch and use it for classification. 
However, as the receptive field of a  deep neural network  at the deep layers is typically larger than the image size, we argue that it is not necessary to crop a patch from the testing images.  

Instead, we propose to pad the testing image such that all the image features can be captured by the model.   Different from directly feeding the image into the model, padding the image with zeros can align the informative region of an image to the center of the receptive filed of the model, similar to random cropping, while getting rid of information loss. Thus it can  further improve the model accuracy.  We name this padding method as \textit{receptive field calibration} (RFC).  RFC can be easily implemented by setting the cropping ratio to be larger than 1, which we call it \textit{RFC ratio},  during the testing phase. 

To investigate its effectiveness, among all the models whose pre-trained weights are publicly available, we apply RFC on the top 2 accurate CNN models  and ViTs respectively. The results are summarized in Tab.~\ref{tab:rfc}. It is observed that RFC can easily further improve the performance of current SOTA model  by 0.11\% \textit{without} fine-tuning.  It is worth noting that the selected models are pre-trained with large auxiliary datasets such as ImageNet-22k and JFT-300M and the performance is nearly saturated. The testing image resolutions   vary from 384 to 800. Under a variety of model configurations, RFC consistently improve the accuracy. With RFC, our proposed Refined-ViT achieves a new SOTA on the ImageNet (among models with less than 100M parameters). 

It should be noted that in the above experiments (Sec.~\ref{exp:analysis} and~\ref{exp:vision_sota}), including comparing Refined-ViT with SOTAs,  we did \textit{not} apply   RFC   for fair comparisons. Here we would like to share such RFC technique as an interesting finding from our recent experiments. RFC is a generic and convenient technique to improve model's classification performance. We believe it is inspiring for rethinking the common random cropping techniques. More importantly, it motivates future studies to dig into the effect  of the model's receptive field   in  classification performance and explore how to calibrate it w.r.t.\ the input images to gain better performance. 

\begin{table*}[h]
\caption{\small RFC can improve both CNN and ViT-based SOTA models on ImageNet, outperforming the strategy of random cropping. We apply RFC on the best performing     pre-trained models   available online.   RFC improves the top-1 accuracy consistently across a wide spectrum of models and configurations, and  establishes  the new SOTA on ImageNet. }
\vspace{-2mm}
\label{tab:rfc}
\footnotesize
\begin{center}
\small
\begin{tabular}{l | c c  | c c  }
\toprule
 Model  & Rand Crop. Ratio &  Top-1 (\%)  & RFC Ratio & Top-1 (\%) \\ [0.5ex] 
\midrule
EfficientNet-l2-ns-800 \cite{tan2019efficientnet} & 0.960 & 88.35 & 1.120 & 88.46  \\ 
EfficientNet-b7-ns-475 \cite{tan2019efficientnet} & 0.936 & 88.23 & 1.020 & 88.33  \\ 
Swin-large-patch4-w12-384 \cite{liu2021swin}  & 1.000 & 87.15 & 1.100 & 87.18  \\ 
% tf-efficientnet-b7-ns \cite{tan2019efficientnet} & 0.949 & 86.8 & 57.4 & 88.4  \\ 
CaiT-m48-448 \cite{gong2021improve} & 1.000 & 86.48 & 1.130 & 86.56  \\ 
\midrule
Refined-ViT-448 & 1.000 & 85.94 & 1.130 & 85.98  \\
\bottomrule

\end{tabular}

\end{center}
\vspace{-2mm}
\end{table*}

\subsection{Applied  to NLP tasks}
We also evaluate the performance of Refiner-ViT models for natural language processing tasks on the GLUE benchmark, to investigate whether Refiner also improves other transformer-based models. We use the BERT-small~\cite{clark2020electra} as the baseline model and replace the self-attention module with refiner, using the same pre-training dataset and recipes. From the results in Tab.~\ref{tab:gluetest}, Refiner boosts the model performance across all the tasks significantly and increases the average score by 1\%, demonstrating Refiner is well generalizable to transformer-based NLP models to improve their attentions and final performance. 

\begin{table*}[h]
\caption{\small Comparison of BERT-small w/o and w/ refiner   on the GLUE development set. }
\vspace{-2mm}
\label{tab:gluetest}
\footnotesize
\begin{center}
\small
\begin{tabular}{l c c c c c c c c c c c}
\toprule
 Model  & Params & MNLI & QNLI & QQP & RTE & SST & MRPC & CoLA & STS-B & Avg.\\ [0.5ex] 
\midrule
BERT-small \cite{clark2020electra}  & 14M & 75.8 & 83.7 & 86.8 & 57.4 & 88.4 & 83.8 & 41.6 & 83.6 & 75.1 \\ 
\quad + Refiner   & 14M & \textbf{78.1} & \textbf{86.4} & \textbf{88.0} & \textbf{57.6} & \textbf{88.8} & \textbf{84.1} & \textbf{42.2} & \textbf{84.0} & \textbf{76.1}\\
\bottomrule

\end{tabular}

\end{center}
\vspace{-2mm}
\end{table*}

\begin{table}[h]
    \centering
    \caption{\small Top-1 accuracy comparison with other methods on ImageNet \cite{deng2009imagenet}
    and ImageNet Real~\cite{beyer2020we}. All models are trained without external data. 
    With the same computation and parameter constraint, Refined-ViT consistently outperforms
    other SOTA CNN-based and ViT-based models. The results of CNNs and ViT are adopted from~\cite{touvron2021going}.}
    \label{tab:sota}
    \def \mysp {\hspace{7pt}}
    \centering \scalebox{1.0}
    {\small 
    \begin{tabular}{@{\ }lccccccc}
    \toprule
    Network  & Params & FLOPs & Train size & Test size  &  Top-1(\%)  & Real Top-1 (\%) \\
    \toprule
    \multicolumn{7}{c}{CNNs}\\
    \midrule

    EfficientNet-B5~\cite{tan2019efficientnet}    & \pzo30M & \dzo9.9B  & $456$ & $456$  & 83.6 & 88.3  \\
    EfficientNet-B7~\cite{tan2019efficientnet}    & \pzo66M & \pzo37.0B & $600$ & $600$  & 84.3 & \_      \\
    Fix-EfficientNet-B8~\cite{tan2019efficientnet, touvron2019fixing} & \pzo87M & \pzo89.5B & $672$ & $800$  & 85.7 & 90.0  \\
    \midrule
    NFNet-F0~\cite{brock2021high}           & \pzo72M & \pzo12.4B & $192$ & $256$ & 83.6 & 88.1  \\
    NFNet-F1~\cite{brock2021high}           & 133M    & \pzo35.5B & $224$ & $320$  & 84.7 & 88.9  \\
    % NFNet-F2~\cite{brock2021high}           & 194M    & \pzo62.6B & $256$ & $352$  & 85.1 & 88.9  \\
    % NFNet-F3~\cite{brock2021high}           & 255M    & 114.8B    & $320$ & $416$  & 85.7 & 89.4  \\
    % NFNet-F4~\cite{brock2021high}           & 316M    & 215.3B    & $384$ & $512$   & 85.9 & 89.4  \\
    % NFNet-F5~\cite{brock2021high}           & 377M    & 289.8B    & $416$ & $544$   & 86.0 & 89.2  \\
    \toprule
    \multicolumn{7}{c}{Transformers}\\
    \midrule
    ViT-B/16~\cite{dosovitskiy2020image}           & \pzo86M & \pzo55.4B & $224$ & $384$ & 77.9 & 83.6 \\
    ViT-L/16~\cite{dosovitskiy2020image}           & 307M    & 190.7B    & $224$ & $384$  & 76.5 & 82.2 \\
    \midrule
    T2T-ViT-14~\cite{yuan2021tokens}       & \pzo22M & \dzo5.2B  & $224$ & $224$        & 81.5 & \_  \\
    T2T-ViT-14$\uparrow$384~\cite{yuan2021tokens} & \pzo22M & \pzo 17.1B  & $224$ & $384$        & 83.3 & \_ \\
    \midrule
    CrossViT~\cite{chen2021crossvit}           & \pzo45M & \pzo56.6B & $224$ & $480$        & 84.1 & \_ \\
    Swin-B~\cite{liu2021swin}             & \pzo88M & \pzo47.0B   & $224$ & $384$        & 84.2 & \_ \\
    TNT-B~\cite{han2021transformer}              & \pzo66M & \pzo14.1B & $224$ & $224$       & 82.8 & \_   \\
    \midrule
    DeepViT-S~\cite{zhou2021deepvit}          & \pzo27M & \dzo6.2B & $224$ & $224$       & 82.3 & \_   \\
    DeepViT-L~\cite{zhou2021deepvit}          & \pzo55M & \pzo12.5B & $224$ & $224$       & 83.1 & \_   \\         
    \midrule
    DeiT-S~\cite{touvron2020training}             & \pzo22M & \dzo4.6B  & $224$ & $224$  & 79.9  & 85.7  \\
    DeiT-B~\cite{touvron2020training}              & \pzo86M & \pzo17.5B & $224$ & $224$ &  81.8 &   86.7 \\
    DeiT-B$\uparrow$384~\cite{touvron2020training}          & \pzo86M & \pzo55.4B & $224$  &  $384$ &  83.1 & 87.7 \\
    \midrule
    CaiT-S36$\uparrow$384~\cite{touvron2021going}  & \pzo68M & \pzo48.0B &  $224$ &  $384$ &  85.4 & 89.8 \\
    CaiT-M36~\cite{touvron2021going} & 271M & \pzo53.7B &  $224$ &  $224$ &  85.1 & 89.3 \\
    \midrule
    LV-ViT-S~\cite{jiang2021token} & \pzo26M & \dzo6.6B &  $224$ &  $224$  &  83.3 & 88.1 \\
    LV-ViT-M~\cite{jiang2021token} & \pzo56M & \pzo16.0B &  $224$ &  $224$  &  84.0 & 88.4 \\
    \toprule
    
    \multicolumn{7}{c}{Transformer with locality} \\
    \midrule
    LocalViT-S~\cite{li2021localvit} & 22.4M & 4.6B & 224 & 224 & 80.8 & - \\
    LocalViT-PVT~\cite{li2021localvit} & 13.5M & 4.8B & 224 & 224 & 78.2 & - \\
    \midrule
    ConViT-S~\cite{d2021convit} & 27M & 5.4B & 224 & 224 & 81.3 & - \\
    ConViT-S+~\cite{d2021convit} & 48M & 10.0B & 224 & 224 & 82.2 & - \\
    \midrule
    BoTNet-S1-59~\cite{srinivas2021bottleneck} & 33.5M & \dzo7.3B & 224 & 224 & 81.7 & -\\
    BoTNet-S1-110~\cite{srinivas2021bottleneck} & 54.7M & \pzo10.9B & 224 & 224 & 82.8 & - \\
    BoTNet-S1-128~\cite{srinivas2021bottleneck} & 79.1M & \pzo19.3B & 256 & 256 & 84.2 & - \\
    BoTNet-S1-128$\uparrow$384~\cite{srinivas2021bottleneck} & 79.1M & \pzo45.8B & 256 & 384 & 84.7 & - \\ \toprule
    \multicolumn{7}{c}{Our \OURS} \\
    \midrule
    \OURS-S & \pzo25M & 7.2B &  $224$ &  $224$  &  83.6 & 88.3 \\
    \OURS-S$\uparrow$384 & \pzo25M & 24.5B &  $224$ &  $384$  & 84.6  &  88.9 \\
    \OURS-M & \pzo55M & 13.5B &  $224$ &  $224$  & 84.6  & 88.9 \\
    \OURS-M$\uparrow$384  & \pzo55M & 49.2B &  $224$ &  $384$ &  85.6 &  89.3 \\
    \OURS-L & \pzo81M & 19.1B &  $224$ &  $224$ &  84.9 & 89.1 \\
    \OURS-L$\uparrow$384 & \pzo81M & 69.1B &  $224$ &  $384$ &  \textbf{85.7} &  \textbf{89.7} \\
    \OURS-L$\uparrow$448 & \pzo81M & 98.0B &  $224$ &  $448$ &  \textbf{85.9} &  \textbf{90.1} \\
    \bottomrule
    \end{tabular}}
\end{table}

\section{Conclusions}
\label{sec:conclusion}
In this work, we introduced the refiner, a simple scheme that augments the self-attention of ViTs by attention expansion and distributed local attention. We find it  works surprisingly well  for improving performance of vision transformers (ViT). Furthermore, it also improves the performance of NLP transformers (BERT) by a large margin. Though refiner is limited to improving diversity of the self-attention maps, we believe its working mechanism is also inspiring for future works on understanding and further improving the self-attention mechanism. 

Some interesting questions  arise from the empirical observations but we leave them open for future studies. First, traditionally all the elements within the attention maps are normalized to be between 0 and 1. However, the attention expansion does not impose such a constraint while giving quite good  results. It is worthy future studies on the  effects of such   subtraction attention (with negative attention weights)   for feature aggregation. 
Second, the local distributed attention indeed imposes a locality inductive bias which is proven to be effective in enhancing learning efficiency of ViTs from medium-sized datasets (ImageNet). How to automatically learn the suitable inductive biases  would be another interesting direction and some recent works have started to explore \cite{d2021convit}.
%
% Thirdly, different from the traditional practice that sets the cropping ratio to be between 0 and 1, RFC sets the cropping ratio to be larger than 1. We have empirically verify that receptive field calibration could steadily improve the classification accuracy on ImageNet. How to incorporate this during the training phases would be another interesting problem to explore.

%%%%%%%%%%%%%%%%%%%%%%%%%%%%%%%%%%%%%%%%%%%%%%%%%%%%%%%%%%%%
% \section*{Checklist}
\newpage

{
\small
\bibliographystyle{plain}
\bibliography{ref}

\begin{thebibliography}{10}

\bibitem{bello2021lambdanetworks}
Irwan Bello.
\newblock Lambdanetworks: Modeling long-range interactions without attention.
\newblock {\em arXiv preprint arXiv:2102.08602}, 2021.

\bibitem{bentivogli2009fifth}
Luisa Bentivogli, Peter Clark, Ido Dagan, and Danilo Giampiccolo.
\newblock The fifth pascal recognizing textual entailment challenge.
\newblock In {\em TAC}, 2009.

\bibitem{beyer2020we}
Lucas Beyer, Olivier~J H{\'e}naff, Alexander Kolesnikov, Xiaohua Zhai, and
  A{\"a}ron van~den Oord.
\newblock Are we done with imagenet?
\newblock {\em arXiv preprint arXiv:2006.07159}, 2020.

\bibitem{brock2021high}
Andrew Brock, Soham De, Samuel~L Smith, and Karen Simonyan.
\newblock High-performance large-scale image recognition without normalization.
\newblock {\em arXiv preprint arXiv:2102.06171}, 2021.

\bibitem{brown2020language}
Tom~B Brown, Benjamin Mann, Nick Ryder, Melanie Subbiah, Jared Kaplan, Prafulla
  Dhariwal, Arvind Neelakantan, Pranav Shyam, Girish Sastry, Amanda Askell,
  et~al.
\newblock Language models are few-shot learners.
\newblock {\em arXiv preprint arXiv:2005.14165}, 2020.

\bibitem{carion2020end}
Nicolas Carion, Francisco Massa, Gabriel Synnaeve, Nicolas Usunier, Alexander
  Kirillov, and Sergey Zagoruyko.
\newblock End-to-end object detection with transformers.
\newblock {\em arXiv preprint arXiv:2005.12872}, 2020.

\bibitem{cer-etal-2017-semeval}
Daniel Cer, Mona Diab, Eneko Agirre, I{\~n}igo Lopez-Gazpio, and Lucia Specia.
\newblock {S}em{E}val-2017 task 1: Semantic textual similarity multilingual and
  crosslingual focused evaluation.
\newblock In {\em Proceedings of the 11th International Workshop on Semantic
  Evaluation ({S}em{E}val-2017)}, pages 1--14, Vancouver, Canada, August 2017.
  Association for Computational Linguistics.

\bibitem{chen2021crossvit}
Chun-Fu Chen, Quanfu Fan, and Rameswar Panda.
\newblock Crossvit: Cross-attention multi-scale vision transformer for image
  classification.
\newblock {\em arXiv preprint arXiv:2103.14899}, 2021.

\bibitem{chen2020pre}
Hanting Chen, Yunhe Wang, Tianyu Guo, Chang Xu, Yiping Deng, Zhenhua Liu, Siwei
  Ma, Chunjing Xu, Chao Xu, and Wen Gao.
\newblock Pre-trained image processing transformer.
\newblock {\em arXiv preprint arXiv:2012.00364}, 2020.

\bibitem{chen2018quora}
Zihan Chen, Hongbo Zhang, Xiaoji Zhang, and Leqi Zhao.
\newblock Quora question pairs, 2018.

\bibitem{clark2020electra}
Kevin Clark, Minh-Thang Luong, Quoc~V Le, and Christopher~D Manning.
\newblock Electra: Pre-training text encoders as discriminators rather than
  generators.
\newblock {\em arXiv preprint arXiv:2003.10555}, 2020.

\bibitem{dagan2005pascal}
Ido Dagan, Oren Glickman, and Bernardo Magnini.
\newblock The pascal recognising textual entailment challenge.
\newblock In {\em Machine Learning Challenges Workshop}, pages 177--190.
  Springer, 2005.

\bibitem{dai2020up}
Zhigang Dai, Bolun Cai, Yugeng Lin, and Junying Chen.
\newblock Up-detr: Unsupervised pre-training for object detection with
  transformers.
\newblock {\em arXiv preprint arXiv:2011.09094}, 2020.

\bibitem{d2021convit}
St{\'e}phane d'Ascoli, Hugo Touvron, Matthew Leavitt, Ari Morcos, Giulio
  Biroli, and Levent Sagun.
\newblock Convit: Improving vision transformers with soft convolutional
  inductive biases.
\newblock {\em arXiv preprint arXiv:2103.10697}, 2021.

\bibitem{deng2009imagenet}
Jia Deng, Wei Dong, Richard Socher, Li-Jia Li, Kai Li, and Li~Fei-Fei.
\newblock Imagenet: A large-scale hierarchical image database.
\newblock In {\em 2009 IEEE conference on computer vision and pattern
  recognition}, pages 248--255. Ieee, 2009.

\bibitem{devlin2018bert}
Jacob Devlin, Ming-Wei Chang, Kenton Lee, and Kristina Toutanova.
\newblock Bert: Pre-training of deep bidirectional transformers for language
  understanding.
\newblock {\em arXiv preprint arXiv:1810.04805}, 2018.

\bibitem{dolan-brockett-2005-automatically}
William~B. Dolan and Chris Brockett.
\newblock Automatically constructing a corpus of sentential paraphrases.
\newblock In {\em Proceedings of the Third International Workshop on
  Paraphrasing ({IWP}2005)}, 2005.

\bibitem{dong2021attention}
Yihe Dong, Jean-Baptiste Cordonnier, and Andreas Loukas.
\newblock Attention is not all you need: Pure attention loses rank doubly
  exponentially with depth.
\newblock {\em arXiv preprint arXiv:2103.03404}, 2021.

\bibitem{dosovitskiy2020image}
Alexey Dosovitskiy, Lucas Beyer, Alexander Kolesnikov, Dirk Weissenborn,
  Xiaohua Zhai, Thomas Unterthiner, Mostafa Dehghani, Matthias Minderer, Georg
  Heigold, Sylvain Gelly, et~al.
\newblock An image is worth 16x16 words: Transformers for image recognition at
  scale.
\newblock {\em arXiv preprint arXiv:2010.11929}, 2020.

\bibitem{giampiccolo2007third}
Danilo Giampiccolo, Bernardo Magnini, Ido Dagan, and Bill Dolan.
\newblock The third pascal recognizing textual entailment challenge.
\newblock In {\em Proceedings of the ACL-PASCAL workshop on textual entailment
  and paraphrasing}, pages 1--9. Association for Computational Linguistics,
  2007.

\bibitem{gong2021improve}
Chengyue Gong, Dilin Wang, Meng Li, Vikas Chandra, and Qiang Liu.
\newblock Improve vision transformers training by suppressing over-smoothing.
\newblock {\em arXiv preprint arXiv:2104.12753}, 2021.

\bibitem{guo2020pct}
Meng-Hao Guo, Jun-Xiong Cai, Zheng-Ning Liu, Tai-Jiang Mu, Ralph~R Martin, and
  Shi-Min Hu.
\newblock Pct: Point cloud transformer.
\newblock {\em arXiv preprint arXiv:2012.09688}, 2020.

\bibitem{haim2006second}
R~Bar Haim, Ido Dagan, Bill Dolan, Lisa Ferro, Danilo Giampiccolo, Bernardo
  Magnini, and Idan Szpektor.
\newblock The second pascal recognising textual entailment challenge.
\newblock In {\em Proceedings of the Second PASCAL Challenges Workshop on
  Recognising Textual Entailment}, 2006.

\bibitem{han2021transformer}
Kai Han, An~Xiao, Enhua Wu, Jianyuan Guo, Chunjing Xu, and Yunhe Wang.
\newblock Transformer in transformer.
\newblock {\em arXiv preprint arXiv:2103.00112}, 2021.

\bibitem{he2016deep}
Kaiming He, Xiangyu Zhang, Shaoqing Ren, and Jian Sun.
\newblock Deep residual learning for image recognition.
\newblock In {\em Proceedings of the IEEE conference on computer vision and
  pattern recognition}, pages 770--778, 2016.

\bibitem{howard2017mobilenets}
Andrew~G Howard, Menglong Zhu, Bo~Chen, Dmitry Kalenichenko, Weijun Wang,
  Tobias Weyand, Marco Andreetto, and Hartwig Adam.
\newblock Mobilenets: Efficient convolutional neural networks for mobile vision
  applications.
\newblock {\em arXiv preprint arXiv:1704.04861}, 2017.

\bibitem{jiang2021token}
Zihang Jiang, Qibin Hou, Li~Yuan, Daquan Zhou, Xiaojie Jin, Anran Wang, and
  Jiashi Feng.
\newblock Token labeling: Training a 85.4\% top-1 accuracy vision transformer
  with 56m parameters on imagenet.
\newblock {\em arXiv preprint arXiv:2104.10858}, 2021.

\bibitem{kornblith2019similarity}
Simon Kornblith, Mohammad Norouzi, Honglak Lee, and Geoffrey Hinton.
\newblock Similarity of neural network representations revisited.
\newblock In {\em International Conference on Machine Learning}, pages
  3519--3529. PMLR, 2019.

\bibitem{lan2019albert}
Zhenzhong Lan, Mingda Chen, Sebastian Goodman, Kevin Gimpel, Piyush Sharma, and
  Radu Soricut.
\newblock {ALBERT}: A lite bert for self-supervised learning of language
  representations.
\newblock {\em arXiv preprint arXiv:1909.11942}, 2019.

\bibitem{levesque2012winograd}
Hector Levesque, Ernest Davis, and Leora Morgenstern.
\newblock The winograd schema challenge.
\newblock In {\em Thirteenth International Conference on the Principles of
  Knowledge Representation and Reasoning}, 2012.

\bibitem{li2017scale}
Jianan Li, Xiaodan Liang, ShengMei Shen, Tingfa Xu, Jiashi Feng, and Shuicheng
  Yan.
\newblock Scale-aware fast r-cnn for pedestrian detection.
\newblock {\em IEEE transactions on Multimedia}, 20(4):985--996, 2017.

\bibitem{li2021localvit}
Yawei Li, Kai Zhang, Jiezhang Cao, Radu Timofte, and Luc Van~Gool.
\newblock Localvit: Bringing locality to vision transformers.
\newblock {\em arXiv preprint arXiv:2104.05707}, 2021.

\bibitem{liu2016recurrent}
Pengfei Liu, Xipeng Qiu, and Xuanjing Huang.
\newblock Recurrent neural network for text classification with multi-task
  learning.
\newblock {\em arXiv preprint arXiv:1605.05101}, 2016.

\bibitem{liu2021swin}
Ze~Liu, Yutong Lin, Yue Cao, Han Hu, Yixuan Wei, Zheng Zhang, Stephen Lin, and
  Baining Guo.
\newblock Swin transformer: Hierarchical vision transformer using shifted
  windows.
\newblock {\em arXiv preprint arXiv:2103.14030}, 2021.

\bibitem{luo2017understanding}
Wenjie Luo, Yujia Li, Raquel Urtasun, and Richard Zemel.
\newblock Understanding the effective receptive field in deep convolutional
  neural networks.
\newblock {\em arXiv preprint arXiv:1701.04128}, 2017.

\bibitem{paszke2019pytorch}
Adam Paszke, Sam Gross, Francisco Massa, Adam Lerer, James Bradbury, Gregory
  Chanan, Trevor Killeen, Zeming Lin, Natalia Gimelshein, Luca Antiga, et~al.
\newblock Pytorch: An imperative style, high-performance deep learning library.
\newblock In {\em Advances in neural information processing systems}, pages
  8026--8037, 2019.

\bibitem{peng2021conformer}
Zhiliang Peng, Wei Huang, Shanzhi Gu, Lingxi Xie, Yaowei Wang, Jianbin Jiao,
  and Qixiang Ye.
\newblock Conformer: Local features coupling global representations for visual
  recognition.
\newblock {\em arXiv preprint arXiv:2105.03889}, 2021.

\bibitem{radford2018improving}
Alec Radford, Karthik Narasimhan, Tim Salimans, and Ilya Sutskever.
\newblock Improving language understanding by generative pre-training, 2018.

\bibitem{rajpurkar-etal-2016-squad}
Pranav Rajpurkar, Jian Zhang, Konstantin Lopyrev, and Percy Liang.
\newblock {SQ}u{AD}: 100,000+ questions for machine comprehension of text.
\newblock In {\em Proceedings of the 2016 Conference on Empirical Methods in
  Natural Language Processing}, pages 2383--2392, Austin, Texas, November 2016.
  Association for Computational Linguistics.

\bibitem{sak2014long}
Hasim Sak, Andrew~W Senior, and Fran{\c{c}}oise Beaufays.
\newblock Long short-term memory recurrent neural network architectures for
  large scale acoustic modeling.
\newblock 2014.

\bibitem{sandler2018mobilenetv2}
Mark Sandler, Andrew Howard, Menglong Zhu, Andrey Zhmoginov, and Liang-Chieh
  Chen.
\newblock Mobilenetv2: Inverted residuals and linear bottlenecks.
\newblock In {\em Proceedings of the IEEE conference on computer vision and
  pattern recognition}, pages 4510--4520, 2018.

\bibitem{shazeer2020talking}
Noam Shazeer, Zhenzhong Lan, Youlong Cheng, Nan Ding, and Le~Hou.
\newblock Talking-heads attention.
\newblock {\em arXiv preprint arXiv:2003.02436}, 2020.

\bibitem{socher2013recursive}
Richard Socher, Alex Perelygin, Jean Wu, Jason Chuang, Christopher~D Manning,
  Andrew~Y Ng, and Christopher Potts.
\newblock Recursive deep models for semantic compositionality over a sentiment
  treebank.
\newblock In {\em Proceedings of the 2013 conference on empirical methods in
  natural language processing}, pages 1631--1642, 2013.

\bibitem{sohn2012learning}
Kihyuk Sohn and Honglak Lee.
\newblock Learning invariant representations with local transformations.
\newblock {\em arXiv preprint arXiv:1206.6418}, 2012.

\bibitem{srinivas2021bottleneck}
Aravind Srinivas, Tsung-Yi Lin, Niki Parmar, Jonathon Shlens, Pieter Abbeel,
  and Ashish Vaswani.
\newblock Bottleneck transformers for visual recognition.
\newblock {\em arXiv preprint arXiv:2101.11605}, 2021.

\bibitem{tan2019efficientnet}
Mingxing Tan and Quoc~V Le.
\newblock Efficientnet: Rethinking model scaling for convolutional neural
  networks.
\newblock {\em arXiv preprint arXiv:1905.11946}, 2019.

\bibitem{tay2020synthesizer}
Yi~Tay, Dara Bahri, Donald Metzler, Da-Cheng Juan, Zhe Zhao, and Che Zheng.
\newblock Synthesizer: Rethinking self-attention in transformer models.
\newblock {\em arXiv preprint arXiv:2005.00743}, 2020.

\bibitem{touvron2020training}
Hugo Touvron, Matthieu Cord, Matthijs Douze, Francisco Massa, Alexandre
  Sablayrolles, and Herv{\'e} J{\'e}gou.
\newblock Training data-efficient image transformers \& distillation through
  attention.
\newblock {\em arXiv preprint arXiv:2012.12877}, 2020.

\bibitem{touvron2021going}
Hugo Touvron, Matthieu Cord, Alexandre Sablayrolles, Gabriel Synnaeve, and
  Herv{\'e} J{\'e}gou.
\newblock Going deeper with image transformers.
\newblock {\em arXiv preprint arXiv:2103.17239}, 2021.

\bibitem{touvron2019fixing}
Hugo Touvron, Andrea Vedaldi, Matthijs Douze, and Herv{\'e} J{\'e}gou.
\newblock Fixing the train-test resolution discrepancy.
\newblock {\em arXiv preprint arXiv:1906.06423}, 2019.

\bibitem{vaswani2017attention}
Ashish Vaswani, Noam Shazeer, Niki Parmar, Jakob Uszkoreit, Llion Jones,
  Aidan~N Gomez, {\L}ukasz Kaiser, and Illia Polosukhin.
\newblock Attention is all you need.
\newblock {\em Advances in neural information processing systems},
  30:5998--6008, 2017.

\bibitem{wang2018glue}
Alex Wang, Amanpreet Singh, Julian Michael, Felix Hill, Omer Levy, and Samuel~R
  Bowman.
\newblock Glue: A multi-task benchmark and analysis platform for natural
  language understanding.
\newblock {\em arXiv preprint arXiv:1804.07461}, 2018.

\bibitem{wang2017residual}
Fei Wang, Mengqing Jiang, Chen Qian, Shuo Yang, Cheng Li, Honggang Zhang,
  Xiaogang Wang, and Xiaoou Tang.
\newblock Residual attention network for image classification.
\newblock In {\em Proceedings of the IEEE conference on computer vision and
  pattern recognition}, pages 3156--3164, 2017.

\bibitem{wang2020end}
Yuqing Wang, Zhaoliang Xu, Xinlong Wang, Chunhua Shen, Baoshan Cheng, Hao Shen,
  and Huaxia Xia.
\newblock End-to-end video instance segmentation with transformers.
\newblock {\em arXiv preprint arXiv:2011.14503}, 2020.

\bibitem{warstadt2019neural}
Alex Warstadt, Amanpreet Singh, and Samuel~R Bowman.
\newblock Neural network acceptability judgments.
\newblock {\em Transactions of the Association for Computational Linguistics},
  7:625--641, 2019.

\bibitem{rw2019timm}
Ross Wightman.
\newblock Pytorch image models.
\newblock \url{https://github.com/rwightman/pytorch-image-models}, 2019.

\bibitem{williams-etal-2018-broad}
Adina Williams, Nikita Nangia, and Samuel Bowman.
\newblock A broad-coverage challenge corpus for sentence understanding through
  inference.
\newblock In {\em Proceedings of the 2018 Conference of the North {A}merican
  Chapter of the Association for Computational Linguistics: Human Language
  Technologies, Volume 1 (Long Papers)}, pages 1112--1122, New Orleans,
  Louisiana, June 2018. Association for Computational Linguistics.

\bibitem{xie2017aggregated}
Saining Xie, Ross Girshick, Piotr Doll{\'a}r, Zhuowen Tu, and Kaiming He.
\newblock Aggregated residual transformations for deep neural networks.
\newblock In {\em Proceedings of the IEEE conference on computer vision and
  pattern recognition}, pages 1492--1500, 2017.

\bibitem{yang2020learning}
Fuzhi Yang, Huan Yang, Jianlong Fu, Hongtao Lu, and Baining Guo.
\newblock Learning texture transformer network for image super-resolution.
\newblock In {\em Proceedings of the IEEE/CVF Conference on Computer Vision and
  Pattern Recognition}, pages 5791--5800, 2020.

\bibitem{yuan2021tokens}
Li~Yuan, Yunpeng Chen, Tao Wang, Weihao Yu, Yujun Shi, Francis~EH Tay, Jiashi
  Feng, and Shuicheng Yan.
\newblock Tokens-to-token vit: Training vision transformers from scratch on
  imagenet.
\newblock {\em arXiv preprint arXiv:2101.11986}, 2021.

\bibitem{yuan2020revisiting}
Li~Yuan, Francis~EH Tay, Guilin Li, Tao Wang, and Jiashi Feng.
\newblock Revisiting knowledge distillation via label smoothing regularization.
\newblock In {\em Proceedings of the IEEE/CVF Conference on Computer Vision and
  Pattern Recognition}, pages 3903--3911, 2020.

\bibitem{zeng2020learning}
Yanhong Zeng, Jianlong Fu, and Hongyang Chao.
\newblock Learning joint spatial-temporal transformations for video inpainting.
\newblock In {\em European Conference on Computer Vision}, pages 528--543.
  Springer, 2020.

\bibitem{zhao2020point}
Hengshuang Zhao, Li~Jiang, Jiaya Jia, Philip Torr, and Vladlen Koltun.
\newblock Point transformer.
\newblock {\em arXiv preprint arXiv:2012.09164}, 2020.

\bibitem{zheng2020end}
Minghang Zheng, Peng Gao, Xiaogang Wang, Hongsheng Li, and Hao Dong.
\newblock End-to-end object detection with adaptive clustering transformer.
\newblock {\em arXiv preprint arXiv:2011.09315}, 2020.

\bibitem{zhou2019neural}
Daquan Zhou, Xiaojie Jin, Qibin Hou, Kaixin Wang, Jianchao Yang, and Jiashi
  Feng.
\newblock Neural epitome search for architecture-agnostic network compression.
\newblock In {\em International Conference on Learning Representations}, 2019.

\bibitem{zhou2021deepvit}
Daquan Zhou, Bingyi Kang, Xiaojie Jin, Linjie Yang, Xiaochen Lian, Qibin Hou,
  and Jiashi Feng.
\newblock Deepvit: Towards deeper vision transformer.
\newblock {\em arXiv preprint arXiv:2103.11886}, 2021.

\bibitem{zhou2018end}
Luowei Zhou, Yingbo Zhou, Jason~J Corso, Richard Socher, and Caiming Xiong.
\newblock End-to-end dense video captioning with masked transformer.
\newblock In {\em Proceedings of the IEEE Conference on Computer Vision and
  Pattern Recognition}, pages 8739--8748, 2018.

\bibitem{zhu2020deformable}
Xizhou Zhu, Weijie Su, Lewei Lu, Bin Li, Xiaogang Wang, and Jifeng Dai.
\newblock Deformable detr: Deformable transformers for end-to-end object
  detection.
\newblock {\em arXiv preprint arXiv:2010.04159}, 2020.

\end{thebibliography}
}

% \input{checklist}

%%%%%%%%%%%%%%%%%%%%%%%%%%%%%%%%%%%%%%%%%%%%%%%%%%%%%%%%%%%%
\newpage

\appendix
% \section{Appendix}
\section{More implementation details}
\subsection{Training hyper-parameters}
For all the ablation experiments, we use the standard training recipe \cite{rw2019timm} that is used for reproducing the ViT baselines. When comparing with other state-of-the-art (SOTA) methods, we adopt the advanced training recipe 
%\footnote{The relabelling method is also used in the same manner as proposed in \cite{jiang2021token}} 
as proposed in \cite{jiang2021token}. Their detailed configurations are shown in Tab. \ref{tab:training_hparam}. We train all models with 8 NVIDIA Telsa-V100 GPUs. 
%\textcolor{red}{[We trained all the models with xx GPUs.]}  
\begin{table}[h]
  \centering
  \footnotesize
  \caption{\small Default training hyper-parameters for our experiments.}
  \label{tab:training_hparam}
  \begin{tabular}{lrr} \toprule[0.5pt]
    H-param. & Standard & Advanced  \\ \midrule[0.5pt] % \midrule[0.5pt]
    Epoch & 300 & 300 \\ \midrule[0.5pt] 
    Batch size & 256 & 512 \\
    LR & 5e-3 $\cdot \frac{\text{batch\_size}}{\text{256}}$ &  5e-3$\cdot \frac{\text{batch\_size}}{\text{512}}$  \\
    LR decay & cosine & cosine\\
    Weight decay & 0.05& 0.05\\
    Warmup epochs& 5 & 5 \\\midrule[0.5pt]
    Dropout & 0 & 0\\
    Stoch. Depth & 0.1 & 0.1 $\cdot \frac{\text{\#Blocks}}{\text{12}}$  \\
    MixUp & 0.2 &  0.8 \\
    CutMix & 0 & 1.0 \\
    Erasing prob. & 0.25 & 0.25 \\
    RandAug & 9/0.5 & 9/0.5\\
    \bottomrule[0.5pt]
  \end{tabular}
\end{table}

\subsection{Fine-tuning with larger image resolutions}
On ImageNet, all models are trained from scratch  with image resolution of $224 \times 224$. When compare with other SOTA models, we fine-tune the models with larger image resolutions using the same fine-tuning procedures as adopted in \cite{touvron2020training}. As larger image resolutions have more positional embeddings, we interpolate the positional encodings in the same manner as the one proposed in \cite{dosovitskiy2020image}. 

\subsection{Model configurations of Refiner-ViT}
As discussed in \cite{yuan2021tokens}, a deep and narrow neural network typically is more efficient  than a wide and shallow one. To verify this, we conduct a pair of experiments using these two architecture configurations respectively and the results are shown in Tab. \ref{tab:arch_effect}. It is clearly observed that with comparable classification accuracy, a deep and narrow network takes 71\% less number of parameters.

\begin{table}[h]
\caption{Comparison of the deep-narrow and shallow-wide architecture configurations.} 
% \vspace{-2mm}
\label{tab:arch_effect}
\footnotesize
\setlength\tabcolsep{2.1pt}
\begin{center}
 \begin{tabular}{l c c c c c c c} 
\toprule
 Model & \#Blocks & Hidden dim & \#Heads   & Params &  Top-1 (\%)\\ 
 \midrule
  \OURS&  12 & 768 & 12 &  86M & 83.1\\ 
  \OURS & 16  & 384 & 12   & 25M & 83.0\\ 
\bottomrule
\end{tabular}
% \vspace{-4mm}
\end{center}
\end{table}

Thus, we design the architecture of Refiner-ViT based on the same deep and narrow network architecture. The detailed configurations and the comparison with the ViT-base architecture are shown in Tab. \ref{tab:arch_config}.

% \begin{table}[h]
% \caption{Model architecture configurations.}
% % \vspace{-2mm}
% \label{tab:arch_config}
% \footnotesize
% \setlength\tabcolsep{2.1pt}
% \begin{center}
%  \begin{tabular}{l | c c c c | c c} 
% \toprule
%  Model & \#Blocks & Hidden dim & \#Head   & \#Params & \#Shared Blocks  & Training Resolution \\ 
%  \midrule
%   \OURS-Base &  12 & 768 & 12 &  86M & 0 & 224 \\ 
%  \midrule
%   \OURS-S & 16  & 384 & 12   & 25M & 1  & 224 \\ 
%   \OURS-M & 32  & 420 & 12   & 55M & 1  & 224  \\ 
%   \OURS-L & 32  & 512 & 12   & 81M & 1  & 224 \\ 
% \bottomrule
% \end{tabular}
% % \vspace{-4mm}
% \end{center}
% \end{table}

\begin{table}[h]
\caption{Model architecture configurations.}
% \vspace{-2mm}
\label{tab:arch_config}
\footnotesize
\setlength\tabcolsep{2.1pt}
\begin{center}
 \begin{tabular}{l | c c c c  c} 
\toprule
 Model & \#Blocks & Hidden dim & \#Head   & \#Params   & Training Resolution \\ 
 \midrule
  \OURS-Base &  12 & 768 & 12 &  86M  & 224 \\ 
 \midrule
  \OURS-S & 16  & 384 & 12   & 25M   & 224 \\ 
  \OURS-M & 32  & 420 & 12   & 55M  & 224  \\ 
  \OURS-L & 32  & 512 & 16   & 81M  & 224 \\ 
\bottomrule
\end{tabular}
% \vspace{-4mm}
\end{center}
\end{table}

\subsection{Feature evolving speed and feature similarity with CKA}
As shown in Fig. 2 and Fig. 4 in the main paper, we compare the similarity between the intermediate token features at the output of each block and the final layer, normalized by the similarity between  the first layer's output features and the final output features as shown in Eqn. (\ref{eqn:cak_similarity}):
\begin{equation}
\label{eqn:cak_similarity}
     F_k  = \frac{\textit{CKA}(f_k, f_{out})}{\textit{CKA}(f_{in}, f_{out})},
\end{equation}
where $f_k$ denotes the token features at layer $k$,  
% and $\textit{CKA}(f_a, f_b)$ denotes the CKA similarity between $f_a$ and $f_b$. 
$f_{in}$ denotes the token features at the first transformer block and $f_{out}$ denotes the features at the final output of the model. The metric $\textit{CKA}$ is calculated with linear kernel function as proposed in \cite{kornblith2019similarity}. The batch size is set to be 32. The similarity score at the output of each block is an average of ten runs with randomly sampled image batches.
% \begin{figure}[!h]
% \centering
% \begin{minipage}{.5\textwidth}
%   \centering
%   \includegraphics[width=.9\linewidth,valign=t]{figures/feature_evolving_fig1.pdf}
%   \captionof{figure}{\small The features of ViT evolves slower than ResNet \cite{he2016deep} and DeiT \cite{touvron2020training} across the model blocks.}
%   \label{fig:feature_baseline}
% \end{minipage}%
% \begin{minipage}{.5\textwidth}
%   \centering
%   \includegraphics[width=.9\linewidth,valign=t]{figures/feature_evolving_conv.pdf}
%   \captionof{figure}{\small Refiner accelerates feature evolving compared with CNNs, the vanilla ViT and the Deit trained with a more complex   scheme.}
%   \label{fig:feature_conv}
% \end{minipage}
% \end{figure}

% \begin{tabular}{p{0.5\textwidth} p{0.5\textwidth}}
%   \vspace{0pt} \includegraphics[width=0.49\textwidth]{figures/feature_evolving_fig1.pdf} &
%   \vspace{0pt} \includegraphics[width=0.49\textwidth]{figures/feature_evolving_fig1.pdf}
% \end{tabular}

\section{More experiments}
 \subsection{Performance improvements of refiner under different training techniques}
 As shown in DeiT~\cite{touvron2020training}, more complex augmentations and fine-tuned training recipe could improve the performance of ViTs significantly without architecture changes. We run a set of experiments to show that the improvements brought by refiner is orthogonal to the training recipes to some extent. We select two training recipes for comparison. The first one uses the searched learning rate schedule and data augmentation policy as proposed in \cite{touvron2020training}. The second one uses  the more complicated patch processing as proposed in \cite{yuan2021tokens}. As shown in Tab.~\ref{tab:tricks}, under different training settings, 
 %although degraded by certain extent, 
 the improvements from the refiner are all significant on ImageNet.
\begin{table}[h]
  \centering
  \small
  \setlength\tabcolsep{1mm}
  \renewcommand\arraystretch{1}
  \caption{Performance gain from refiner under different training recipes.} 
  \label{tab:tricks}
  \begin{tabular}{lcclcc} \toprule[0.5pt]
    Training techniques & \#Param. & Original Top-1  (\%) &  + Refined-ViT \\ \midrule[0.5pt]  
    Baseline (ViT-Base) & 86M &  79.3 & 81.2
    (\textcolor{ForestGreen}{\textbf{+1.9}}) \\
    + DeiT-Base (\cite{touvron2020training}) & 86M & 81.5 & 82.3 (\textcolor{ForestGreen}{\textbf{+0.8}}) \\
    + More convs for patch embedding &86M& 82.2 & 83.1 (\textcolor{ForestGreen}{\textbf{+0.9}})\\
    \bottomrule
  \end{tabular}
\end{table}

\subsection{Impacts of the attention heads}
As discussed in \cite{touvron2021going}, increasing the number of attention heads will reduce the embedding dimension per head and thus degrade the network performance. Differently, refiner expands the number of heads with a linear projection function. With the implementation of the linear projection, refiner will keep each head's embedding dimension unchanged. As shown in Tab. \ref{tab:attn_head_ablation}, the performance keeps increasing with Refined-ViT.
\begin{table}[h]
    \footnotesize
    \setlength\tabcolsep{2.1pt}
    \centering
    \caption{Ablation on the impact of the number of the heads for   ViT-Small with 16 blocks and 384 embedding dimension. We report top-1 accuracy on ImageNet.  The accuracy for ViT-Small with different head numbers are adopted from \cite{touvron2021going}.}
% \vspace{-2mm}
    \label{tab:attn_head_ablation}
        \begin{tabular}{ccccc} \toprule
    \# Heads & ViT-Small  &  Refined-ViT \\ \midrule
     8  & 79.9  & 82.5\\ 
    12 & 80.0  & 82.6\\ 
    16 &  80.0  & 82.8\\
    \bottomrule
    \end{tabular}
% \vspace{-4mm}
\end{table}

\subsection{Impacts of the kernel size}
% \begin{wrapfigure}[11]{r}{0.5\textwidth}
\begin{figure}
    \centering
    % \vspace{-20mm}
    \includegraphics[width=0.5\textwidth]{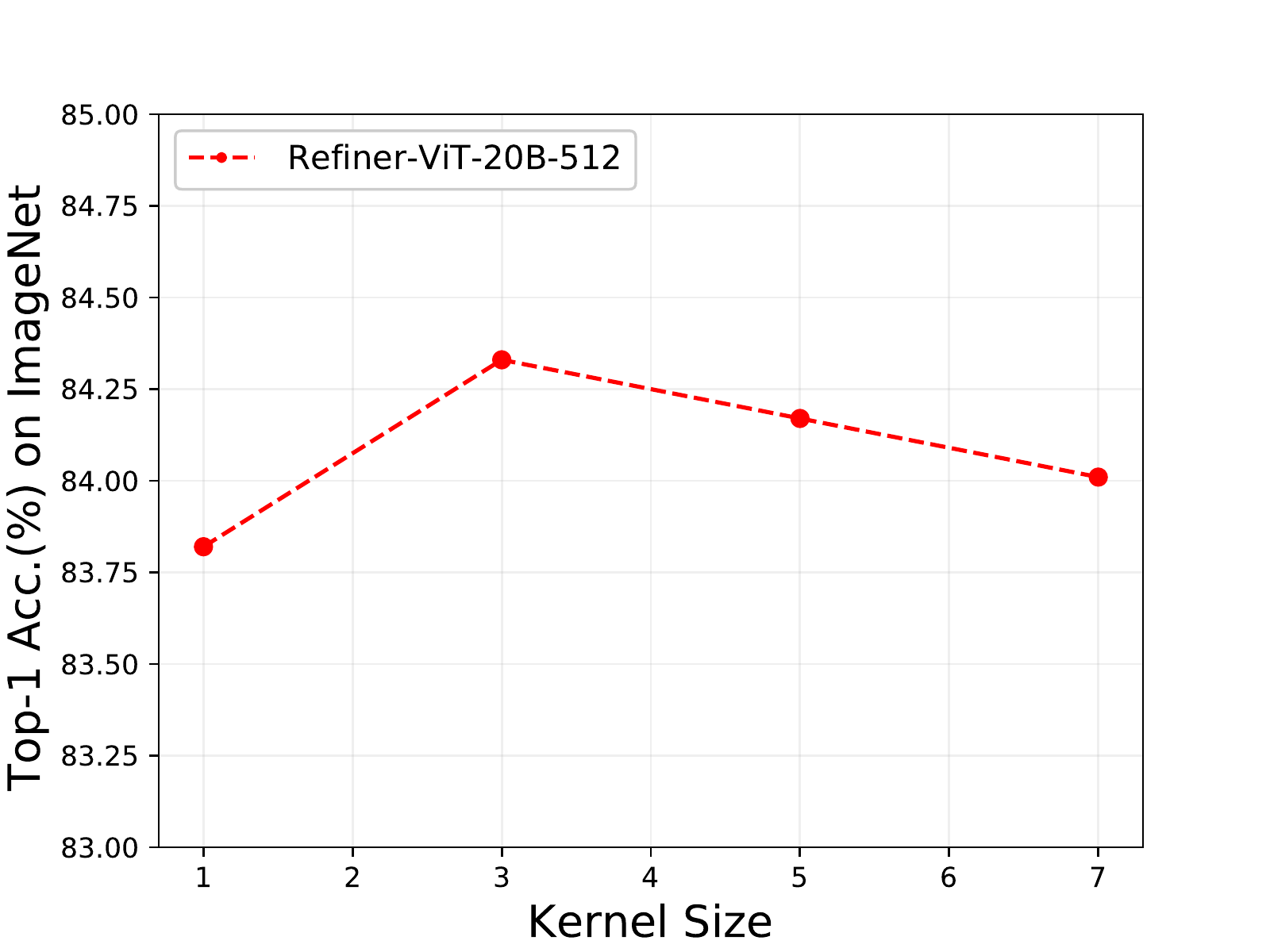}
    % \vspace{-4mm}
    \caption{\small Top-1 classification accuracy of Refined-ViT with different local kernel sizes.}
    \label{fig:kernel_size}
\end{figure}
% \end{wrapfigure}
For all the experiments in the main paper, the convolution kernel is set to be $3 \times 3$ by default. In Fig. \ref{fig:kernel_size}, we show the classification results on ImageNet with different kernel size.  It is observed that all kernels achieve much better accuracy than the baseline method (kernel size of 1) and the refiner with kernel size 3 achieves that highest accuracy. This further verify that attention reinforcement within the local area could improve the performance of ViTs. 
% We argue the selection of the kernel size is related to the length of the token sequence. As we have 20 transformer blocks in the experiments, the convolution with kernel size 3 and 5 have receptive fields of 95 and 175 receptively. Further increase the kernel size will cause the receptive field to be larger than the length of the token sequence (197) and thus degrade the performance.

\subsection{Refiner makes the attention maps shareable} 
% \begin{wraptable}[10]{r}{4cm}
\begin{table}[h]
% \vspace{-7mm}
  \centering
  \small
  \setlength\tabcolsep{1mm}
  \renewcommand\arraystretch{1}
  \caption{\small Ablation on the number of shared attention maps on ViT-16B with 384 embedding dimension.  }
  \label{tab:shared_attn}
  \begin{tabular}{cccc} \toprule[0.5pt]
     \#Shared Blocks  & SA &  Refiner
     \\ \midrule[0.5pt]  
    0 &   82 & 83.0
     \\
     1  &   77 & 83.0 \\
     2 &  70 & 82.8 \\
     3 &  NAN & 82.6 \\
    \bottomrule
  \end{tabular}
\end{table}
% \end{wraptable} 
Refiner enables the attention maps to capture some critical spatial relation among the tokens. 
To see this,  we compute the attention maps with Refiner and let the following a number of blocks to reuse such pre-computed attention maps. The results are given in
Tab.~\ref{tab:shared_attn}.  Interestingly we find that the attention maps after Refiner can be shared (reused) by several following SA blocks. Let the directly adjacent SA block reuse the refined attention map from the previous block will maintain the model overall performance. Sharing the attention maps with the following two SA blocks only drop the accuracy by 0.2\%. In stark contrast, the attention maps from the vanilla SA are completely not shareable, leading the model performance to drop severely. This demonstrates Refiner is effective at extracting useful patterns and this finding is also in accordance with the one from a recent NLP study~\cite{tay2020synthesizer}.  Therefore, across the experiments with Refiner, we shared the attention maps for 1 adjacent transformer block. 

\section{Dataset description for NLP experiments}
% \subsubsection{GLUE dataset}
GLUE benchmark \cite{wang2018glue} is a collection of nine natural language understanding tasks where the labels of the testing set is hidden and the researchers need to submit their predictions to the evaluation server\footnote{\url{https://gluebenchmark.com}} to obtain results on testing sets. In this work, we only present results of single-task setting for fair comparison. The nine tasks included in GLUE benchmark are described in details as below.
\paragraph{MNLI}
The Multi-Genre Natural Language Inference Corpus \cite{williams-etal-2018-broad} is a dataset of sentence pairs with textual entailment annotations. Given a premise sentence
and a hypothesis sentence, the task is to predict their relationships including \textsc{ententailment}, \textsc{contradiction} and  \textsc{neutral}. The data is collected from ten distinct genres including both written and spoken English.

\paragraph{QNLI}
Question Natural Language Inference is a binary sentence pair classification task, converted from The Stanford Question Answering Dataset \cite{rajpurkar-etal-2016-squad} (a question-answering dataset). Each sample of QNLI contains a context sentence and a question. The task is to determine whether the context sentence contains
the answer to the question.
\paragraph{QQP}
The Quora Question Pairs dataset \cite{chen2018quora} is a collection of question pairs from Quora (a community
question-answering website). The task is to determine whether the two questions in a pair are semantically equivalent. 
\paragraph{RTE}
Similar to MNLI, the Recognizing Textual Entailment (RTE) dataset is also a binary classification task, i.e., \textit{entailment} and \textit{not entailment}. It is from a series of annual textual
entailment challenges including RTE1 \cite{dagan2005pascal}, RTE2 \cite{haim2006second}, RTE3 \cite{giampiccolo2007third}, and RTE5 \cite{bentivogli2009fifth}.

\paragraph{SST-2}
The Stanford Sentiment Treebank \cite{socher2013recursive} consists of sentences from movie
reviews and human annotations of their sentiment. GLUE uses the two-way (\textsc{positive/negative}) class split.

\paragraph{MRPC}

The Microsoft Research Paraphrase Corpus \cite{dolan-brockett-2005-automatically} contains data from online news that consists of sentence pairs with human annotations regarding whether the sentences in the pair are semantically equivalent. 

\paragraph{CoLA}

The Corpus of Linguistic Acceptability \cite{warstadt2019neural} is a binary single-sentence classification dataset containing the examples annotated with whether it is a grammatical English sentence.

\paragraph{SST-B}
The Semantic Textual Similarity Benchmark \cite{cer-etal-2017-semeval} is a collection of human-annotated sentence pairs  with a similarity score ranging from 1 to 5. The target is to to predict the scores with a given sentence.

\paragraph{WNLI}
Winograd NLI \cite{levesque2012winograd} is a small natural language inference dataset. However, as highligthed on the GLUE web page\footnote{\url{https://gluebenchmark.com/faq}}, there are issues with the construction of it. Thus following the practice of previous works (GPT \cite{radford2018improving} and BERT \cite{lan2019albert} etc.), we exclude this dataset for a fair comparison.

\end{document}